\pdfoutput=1
\PassOptionsToPackage{dvipsnames}{xcolor}
\PassOptionsToPackage{pdftex}{xcolor}
\documentclass[11pt]{article}
\usepackage[normalem]{ulem}
\usepackage{acl}

\usepackage{color,soul} 
\usepackage{times}
\usepackage{latexsym}
\usepackage{algorithm}
\usepackage{algpseudocode}
\usepackage{times}
\usepackage{latexsym}
\usepackage{smartdiagram}
\usepackage{appendix}
\usepackage{tikz}
\usepackage{tcolorbox}
\usetikzlibrary{arrows.meta,
                positioning
                }
\usepackage{amsmath}
      
\usepackage{hwemoji}
\usepackage{hyperref}
\usepackage{xcolor}

\usepackage{changepage, threeparttable} 

\definecolor{topPerformer}{rgb}{0.82, 0.94, 0.75} 
\definecolor{secondBest}{rgb}{0.91, 0.84, 0.42}  

\usepackage[T1]{fontenc}

\usepackage[utf8]{inputenc}

\usepackage{microtype}
\usepackage{booktabs}
\usepackage{graphicx}

\usepackage{inconsolata}
\usepackage{microtype}
\usepackage{amsmath}
\usepackage{amssymb}

\usepackage{inconsolata}
\usepackage{amsthm, amsmath, amsfonts}
\usepackage{bm}
\usepackage{pgfplots}
\pgfplotsset{compat=1.17}
\definecolor{deepblue}{HTML}{1f77b4}
\definecolor{vibrantorange}{HTML}{ff7f0e}
\definecolor{lushgreen}{HTML}{2ca02c}
\definecolor{elegantpurple}{HTML}{9467bd}
\definecolor{olivegreen}{RGB}{128, 128, 0}
\definecolor{darkgreen}{RGB}{0, 100, 0}
\definecolor{limegreen}{RGB}{50,205,50}
\definecolor{salmon}{RGB}{250,128,114}
\definecolor{mistyrose}{RGB}{255,228,225}
\definecolor{colorA}{RGB}{26,112,149}    
\definecolor{colorB}{RGB}{241,171,64}  
\definecolor{colorC}{RGB}{120,175,92}   
\definecolor{colorD}{RGB}{151,65,113} 
\newcommand{\colorSquare}[1]{\colorbox{#1}{\rule{0pt}{1mm}\rule{1mm}{0pt}}}

\newcommand{\clvo}[1]{{\color{vibrantorange} #1}}
\newcommand{\cllg}[1]{{\color{lushgreen} #1}}

\newcommand{\clr}[1]{{\color{red} #1}}
\newcommand{\clb}[1]{{\color{blue} #1}}

\newcommand{\cldg}[1]{{\color{darkgreen} #1}}
\newcommand{\clm}[1]{{\color{magenta} #1}}
\newcommand{\clo}[1]{{\color{orange} #1}}
\newcommand{\clp}[1]{{\color{purple} #1}}
\tikzset{
  heading/.style={
    rectangle,
    rounded corners,
    draw=gray,
    align=center,
    minimum width=8em,
    text width=8em,
    font=\scriptsize,
    fill=yellow!50!red!20,
    top color=yellow!50!red!30,
    bottom color=yellow!50!red!10,
    drop shadow={shadow xshift=2pt, shadow yshift=-2pt, fill=gray!70, opacity=0.6}
  },
  example_heading/.style={
    rectangle,
    rounded corners,
    draw=gray,
    align=center,
    minimum width=8em,
    text width=10em,
    font=\scriptsize,
  },
  example_prompt/.style={
    rectangle,
    rounded corners,
    draw=gray,
    align=center,
    minimum width=8em,
    text width=40em,
    font=\scriptsize,
  },
  inline_prompt/.style={
    rectangle,
    rounded corners,
    draw=gray,
    align=left,
    minimum width=18em,
    text width=18em,
    font=\scriptsize,
  },
  prompt/.style={
    rectangle,
    rounded corners,
    draw=gray,
    align=left,
    minimum width=18em,
    text width=18em,
    font=\scriptsize,
    fill=green!60!blue!20,
    top color=green!70!blue!30,
    bottom color=green!70!blue!10,
    drop shadow={shadow xshift=2pt, shadow yshift=-2pt, fill=gray!70, opacity=0.6}
  },
  response/.style={
    rectangle,
    rounded corners,
    draw=gray,
    align=center,
    minimum width=18em,
    text width=18em,
    font=\scriptsize,
    fill=yellow!50!red!20,
    top color=yellow!50!red!30,
    bottom color=yellow!50!red!10,
    drop shadow={shadow xshift=2pt, shadow yshift=-2pt, fill=gray!70, opacity=0.6}
  },
  arrow/.style={
    ->,
    draw=gray,
    line width=2mm,
    shorten >=1pt,
    shorten <=1pt,
    -{Triangle[scale=0.6]}
  }
}
\newcommand{\CircledA}[1]{%
    \tikz[baseline=(char.base)]{
        \node[shape=circle,draw,inner sep=2pt, color=blue, fill=green] (char) {#1};}
}
\newcommand{\CircledC}[1]{%
    \tikz[baseline=(char.base)]{
        \node[shape=circle,draw,inner sep=2pt, color=blue, fill=red] (char) {#1};}
}
\usepackage{inconsolata}
\usepackage{arydshln}
\usepackage{multirow}
\usepackage[colorinlistoftodos,prependcaption,textsize=tiny]{todonotes}
\definecolor{olivegreen}{RGB}{107,142,35}
\definecolor{lightolivegreen}{RGB}{157,192,105}

\newcommand{\ourmodel}[1]{FinTral}

\newcommand{\modela}[1]{\textit{$FinBEAT$}}
\newcommand{\modelb}[1]{\textit{$FinBEAT_{v2}$}}
\newcommand{\modelc}[1]{\textit{$FinBEAT_{v3}$}}
\title{\raisebox{-0.35\height}{\includegraphics[height=1cm,width=1cm]{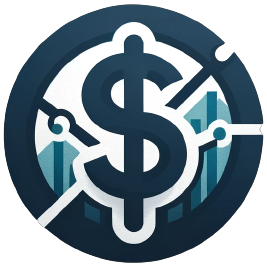}}
\ourmodel~: A Family of GPT-4 Level Multimodal Financial Large Language Models}

\author{ \textbf{Gagan Bhatia }~~~~ \textbf{El Moatez Billah Nagoudi}~~~~ \textbf{ Hasan Cavusoglu}\\ ~~~~~~~~~~~\textbf{Muhammad Abdul-Mageed}~~~~~\\ 
\\The University of British Columbia \&
   Invertible AI \\
   \texttt{\normalsize \{gagan30@student., moatez.nagoudi@, cavusoglu@sauder.\}ubc.ca} \\
\texttt{\normalsize \{muhammad.mageed@\}ubc.ca; invertible.ai}}

\begin{document}
\maketitle
\begin{abstract}
We introduce\textit{~\ourmodel~}, a suite of state-of-the-art multimodal large language models (LLMs) built upon the Mistral-7b model and tailored for financial analysis.~\ourmodel~ integrates textual, numerical, tabular, and image data. We present \textbf{FinSet}, the largest financial LLM pretraining training, instruction tuning and financial alignment dataset and evaluation benchmark featuring nine tasks and 23 datasets and the first to understand and mitigate financial hallucinations. We enhance \textbf{\ourmodel~} with domain-specific pretraining, instruction fine-tuning, and RLAIF training by exploiting a large collection of textual and visual datasets we curate for this work. Our ~\ourmodel~ model trained with direct preference optimization employing advanced \textbf{T}ools and \textbf{R}etrieval methods, dubbed~\textit{\ourmodel~-DPO-T\&R}, demonstrates an exceptional zero-shot performance. It outperforms ChatGPT-3.5 in all tasks and surpasses GPT-4 in five out of nine tasks, marking a significant advancement in AI-driven financial technology. We also demonstrate that~\ourmodel~ has the potential to excel in real-time analysis and decision-making in diverse financial contexts. The GitHub repository for \textit{FinTral} is available at \url{https://github.com/UBC-NLP/fintral}.
\end{abstract}

\section{Introduction}

Natural Language Processing (NLP) plays a key role in financial document analysis, interpretation, and utilization. In recent years, a wide range of applications incorporating advances in NLP have emerged. These include sentiment analysis of financial news, event extraction from financial documents, and the generation and summarization of financial reports~\cite{souma2019enhanced,araci2019finbert,yang2018dcfee}. These developments have uncovered the potential for unstructured data for data-driven financial decision-making and the transformation of financial documents into actionable insights and market intelligence. 
\begin{figure}[t]
\includegraphics[width=\linewidth]{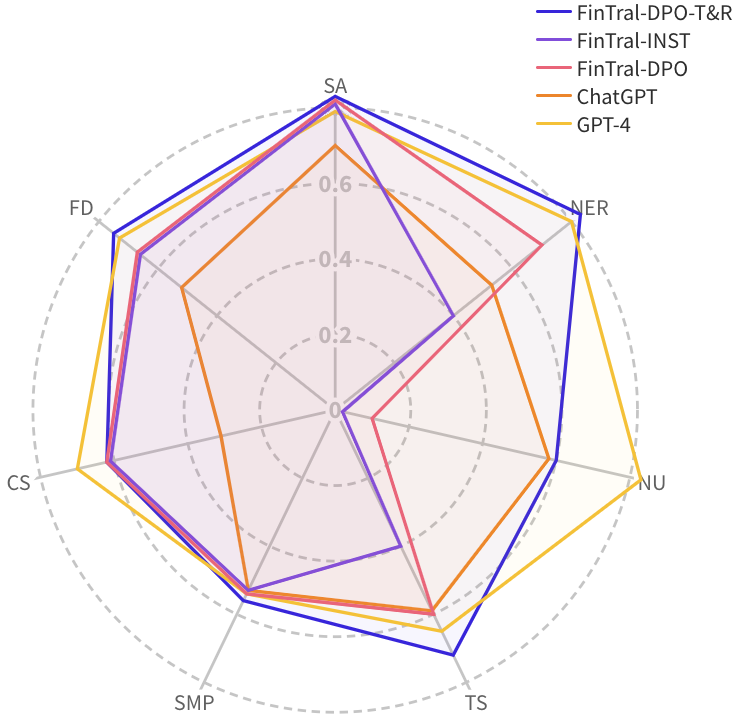}
\centering
\caption{Comparative Performance Analysis on text-based tasks of Key Financial AI Models. We compare three variations of \textbf{\textit{FinTral}} with ChatGPT (GPT-3.5) and GPT-4 across
seven task clusters: Sentiment Analysis (SA), Named Entity Recognition (NER), Number Understanding (NU), Text Summarization (TS), Stock Movement Prediction (SMP), Credit Scoring (CS), and Firm Disclosure (FD).
}
\label{fig:radarplot}
\end{figure} 
Applying NLP in finance, however, is challenging because financial documents often include dense numerical information and domain-specific jargon requiring advanced numerical processing and reasoning capabilities \cite{mik2017smart,liu2023fingpt}. This means that financial NLP models need extensive domain knowledge before they can capture the nuanced implications of accounting and financial measures, economic indicators, and market trends. This is also compounded by the rapid pace of financial markets, where real-time analysis is crucial but challenging to achieve~\cite{gupta2023gptinvestar,yang2023investlm}. 

Similar to other domains, large language models (LLMs) are starting to disrupt financial document understanding ~\cite{chapman2022towards, la2020end} but can also suffer from the same issues as transitional approaches. LLMs are also prone to hallucination, reducing their usability in financial decision-making~\cite{kang2023deficiency}. Financial documents can also involve various types of visual content, which require models with multimodal abilities. 

To meet these challenges, we introduce a groundbreaking LLM specialising in the financial domain. Our model, dubbed\textit{~\ourmodel}, is designed to overcome hurdles of the financial domain through a multimodal approach that integrates textual, numerical, tabular, and visual data processing for comprehensive document understanding. We train our model off Mistral-7b~\cite{jiang2023mistral} on a sizeable domain-specific dataset and instruction-tune it for the financial domain using extensive instruction data. We then carefully align it with GPT-4 generated responses leveraging the recently introduced direct policy optimization (DPO) method \cite{rafailov2023direct}. In order to evaluate~\ourmodel~, we introduce an extensive benchmark of eight different tasks based on 25 different datasets. Our model outperforms all other models of comparable size and, in spite of its much smaller size, performs on par with GPT-4. 


To summarize, we offer the following contributions: \textbf{(1)} We introduce \ourmodel, a cutting-edge multimodal LLM specialized in financial data, and FinSet, an extensive financial LLM training and evaluation benchmark. FinSet is the largest financial evaluation benchmark and the only one that measures model hallucinations, encompassing nine tasks across 25 datasets. \textbf{(2)} \ourmodel is further instruction-finetuned and carefully aligned using the DPO objective, using AI feedback data, resulting in \textit{\ourmodel-DPO}. \textbf{(3)} We have also endowed ~\ourmodel~ with vision capabilities, extending it to \textit{\ourmodel-VL}, which employs the CLIP~\cite{radford2021learning} vision encoder. For enhanced performance, we developed a version that utilizes \textbf{T}ools and \textbf{R}etrieval, \textit{\ourmodel-DPO-T\&R}. \textbf{(4)} \ourmodel-DPO demonstrates exceptional zero-shot capabilities, outperforming ChatGPT~\cite{openai2023chatgpt} in \textit{all} tasks. Moreover, our best model, \ourmodel-DPO-T\&R, surpasses GPT-4~\cite{openai2023gpt4} in five of eight text-based tasks.

\begin{figure*}[ht]
\includegraphics[width=\textwidth]{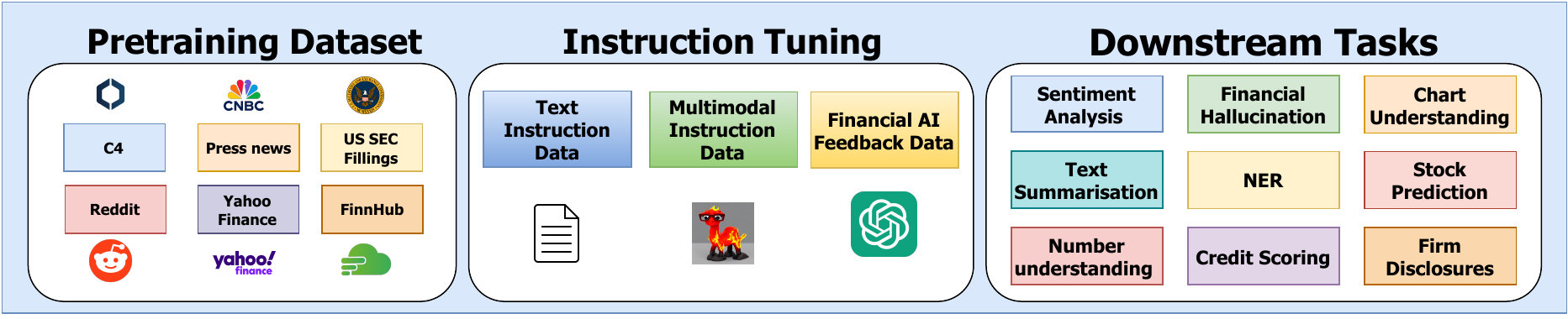}
\centering
\caption{FinSET, a Financial Training and Evaluation Benchmark.} \label{fig:finset}
\end{figure*} 
The rest of this paper is organized as follows: 
In Section~\ref{related_works}, we review related work with a particular emphasis on financial LLMs, their applications and challenges. Section~\ref{data} outlines how we built our benchmark dataset: FinSet. We present our approach to model pretraining, instruction tuning, and prompting strategies, and subsequently introduce FinTral models in Section~\ref{models}. In Section~\ref{results}, we present our experiments and comprehensively analyse our models. We discuss our results in Section \ref{discuss} and conclude in Section~\ref{conclusion}. Finally, we discuss limitations in Section~\ref{sec:limit} and provide an ethics statement in Section~\ref{sec:ethics}.

\section{Related Works} \label{related_works}
\noindent\textbf{NLP for finance}
Traditional NLP has been applied to various finance tasks, including named entity recognition, sentiment analysis, event extraction, financial report generation, and text summarization~\cite{salinas-alvarado-etal-2015-domain, souma2019enhanced,araci2019finbert, yang2018dcfee,zheng2019doc2edag, chapman2022towards, la2020end}. However, traditional models face challenges in this domain due to complexity of financial language, scarcity of annotated data, limited inferential capabilities, and the need for real-time analysis. Adaptability of conventional NLP models is also limited, with such models often optimized for single-task functions~\cite{mik2017smart,mishra2021cross,liu2023fingpt}.

\noindent\textbf{Financial LLMs }
Advancements in financial models began with FinBERT~\cite{araci2019finbert}. Recently, models like BloombergGPT~\cite{wu2023bloomberggpt}, PIXIU~\cite{xie2023pixiu}, Instruct-FinGPT~\cite{zhang2023instructfingpt}, and GPT-FinRE~\cite{rajpoot2023gptfinre} are notable contributions. Other innovations include introduction of multimodal capabilities (FinVis-GPT \cite{wang2023finvisgpt}), enhancement of investment strategies (GPT-InvestAR \cite{gupta2023gptinvestar}, InvestLM \cite{yang2023investlm}), and efforts to address challenges such as economic sentiment analysis and hallucination in information extraction~\cite{zhang2023enhancing, sarmah2023reducing}. FinLMEval~\cite{guo2023chatgpt} and DISC-FinLLM~\cite{chen2023discfinllm} focus on evaluation and model performance in monetary scenarios. Other work, such as \citet{chu2023datacentric}, emphasizes sophisticated data preprocessing for better handling of financial tasks. Appendix~\ref{subsec:literature} provides a further discussion of the NLP and LLMs literature in finance.

\section{FinSet 
} \label{data}


We develop comprehensive and diverse datasets to build~\ourmodel~. We first describe our raw datasets rich in domain-specific tokens, setting a solid foundation for model training, then our instruction finetuning and AI-driven feedback datasets. Subsequently, we present a multi-modal financial dataset to facilitate a nuanced approach to data interpretation. Finally, we introduce an extensive set of evaluation benchmark datasets tailored to test the model's performance across diverse financial tasks.


\subsection{Pretraining Dataset}
\begin{figure*}[ht]
\includegraphics[width=\textwidth]{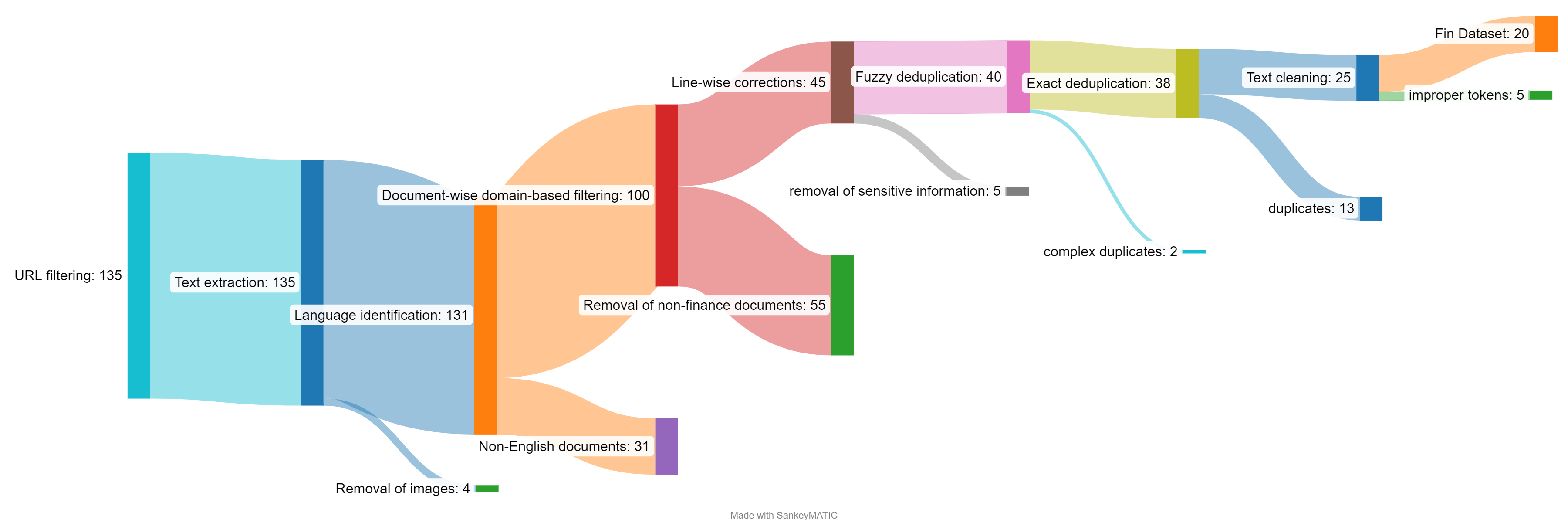}
\centering
\caption{Illustration of our data deduplication pipeline} \label{fig:dedup_pipeline}
\end{figure*}

We introduce FinSet, a 20 billion token, high quality dataset we build for financial LLM training. FinSet is acquired based on a collection of large text corpora (2.9 billion documents, making 135 billion tokens; see Table \ref{tab: raw_data}) from which we extract finance-specific data that we then clean using a careful filtering pipeline. The datasets are described in Appendix~\ref{pretraindata}. Our cleaning pipeline is detailed in Appendix~\ref{dedup} and is further illustrated in \autoref{fig:dedup_pipeline}. 
Our document cutoff date is August 1, 2023, which affords recent information to our models.

\begin{table}[!htp]\centering
\scriptsize
\begin{tabular}{lrrrrr}\toprule
\textbf{Dataset} &\textbf{Documents} &\textbf{Tokens} &\textbf{Deduplicated Tokens} \\\toprule
C4 &2.8B &124.0B &11.75B\\
News &51.5M &8.7B &5.65B\\
SEC &4.3M &3.1B &2.55B \\
Social Media &717.7K &8.2M &7.87M \\
Press &12.0K &3.1M &1.55M \\
\midrule
\textbf{Total} &2.9B &135.9B &\textbf{20.0B} \\
\bottomrule
\end{tabular}
\caption{Details of our pretraining resources. 
}
\label{tab: raw_data}
\end{table}


\subsection{Financial Instruction Data}\label{subsec:instruct_data}
\label{instruction_data}
We assemble an extensive collection of instruction tuning datasets to enhance capabilities of our models. 
The datasets originate from various sources, notably including interactions with GPT-3.5 and GPT-4 for a diverse host of tasks. Again, we apply a deduplication and filtering pipeline (detailed in Appendix \ref{dedup}) to exclude non-financial instructions, thereby focusing solely on financial reasoning. Table \ref{tab: instructions} shows our various data sources, along with the resultant (final) dataset. 

\begin{table}[!ht]
\scriptsize
\begin{tabular}{llr}\toprule
\textbf{Dataset} &\textbf{Source} &\textbf{Instructions} \\\toprule
FLUPE &ChanceFocus/FLUPE &123.0k \\
finance-alpaca & Gbharti/Finance-alpaca &68.91k \\
finest-finred &FinGPT/Hingpt-finred &32.67k \\
Math Instruct &TIGER-Lab/MathInstruct &26.2k \\
fin-llama-dataset &bavest/fin-llama-dataset &16.9k \\
llama-2-finance &AdiOO7/llama-2-finance &4.84k \\\midrule
\multicolumn{1}{l}{Total instructions} &   \multicolumn{1}{c}{-}& 272.6k \\
\multicolumn{1}{l}{\textbf{Total after deduplication} } & \multicolumn{1}{c}{-} &\textbf{226.3k} \\
\bottomrule
\end{tabular}
\caption{Instruction tuning datasets.
}\label{tab: instructions}
\end{table}



\subsection{Financial AI Feedback Data}\label{feedback_data}

Human feedback is valuable for aligning LLMs. Traditionally, this feedback is derived from human preferences as to the quality of LLM responses. In this work, we employ AI feedback through a refined version of the finance reasoning instruction dataset described in Section ~\ref{subsec:instruct_data}. 

Along with the output generated by GPT-4~\cite{gpt4}, we  generate responses using the FinMA-7B~\cite{xie2023pixiu} and LLaMa-7B-chat~\cite{touvron2023llama} models to each prompt. For a given prompt, the GPT-4 output is selected as the `chosen' response while we select randomly one from FinMA-7B and LLaMa outputs as the `rejected' response. 
Our AI feedback data includes a total of 43k samples, and we show an example of this data in Figure ~\ref{listing:dpo_data}.

\subsection{Visual Financial Instruction Dataset}
\noindent For aligning the vision language components in~\ourmodel~, we use LAION, CC, and SBU datasets from the Llava pretraining data~\cite{liu2023improved}. We also use the ChartQA training set~\cite{masry-etal-2022-chartqa} for the same purpose. In addition, we follow the same approach by ~\citet{wang2023finvisgpt} to further expand our visual pretraining dataset. While \citet{wang2023finvisgpt} use Chinese data, we use the Fortune-500 companies stock price data, allowing us to create our own English dataset, dubbed \textit{FinVis-PT}. We then use LLava Instruct data to improve the instruction understanding of our multimodal LLMs, creating our instruction tuning dataset \textit{FinVis-IT}. While the FinVis-PT dataset includes stock market charts and asks simple questions about them, FinVis-IT is multi-turn and includes more complex charts and instructions. Our visual instruction datasets are described in Table~\ref{tab: vl-data}.
\begin{table}[!htp]\centering
\scriptsize
\resizebox{\columnwidth}{!}{%
\begin{tabular}{lllr}\toprule
\textbf{Multimodal Training} &\textbf{Dataset} &\textbf{Source} &\textbf{Instructions} \\\toprule
\multirow{3}{*}{\textbf{Alignment}} &LAION/CC/SBU & \newcite{liu2023improved}  &558k \\
&FinVis-PT & \textbf{Our Paper} &185k \\
&ChartQA & \newcite{masry-etal-2022-chartqa} &20.9k \\\midrule
\multirow{2}{*}{\textbf{Multiturn}} &FinVis-IT &\textbf{Our Paper} &427k \\
&LLava 1.5 & \newcite{liu2023improved} &665k \\
\toprule
\multicolumn{1}{l}{\textbf{Total}}& & &1.1M \\
\bottomrule
\end{tabular}
}
\caption{Visual financial instruction datasets. We generated FinVis using the same method from \citet{wang2023finvisgpt}. }
\label{tab: vl-data}
\end{table}

\subsection{Downstream Evaluation Datasets}\label{donwstream_tasks_data}

\begin{table*}[htb!]
\centering
\scriptsize
\resizebox{\textwidth}{!}{%
\begin{tabular}{llrllll}\toprule
\textbf{Data} &\textbf{Task} &\textbf{Instruction} &\textbf{Data Types} &\textbf{Modalities} &\textbf{Source} &\textbf{Metrics} \\\toprule

ChartQA & \multirow{3}{*}{chart understanding} &$2,500$ &general charts &  \multirow{3}{*}{text, images} &\newcite{masry-etal-2022-chartqa} &\multirow{3}{*}{Accuracy} \\
FinVQAv1 &  &$500$ &stock market charts &  &\textbf{Our paper} &  \\
FinVQAv2 &  &$525$ &complex financial charts &  &\textbf{Our paper} &  \\

\midrule

Australian & &$690$ &  &  &\newcite{misc_statlog_(australian_credit_approval)_143} & \\
German &\multirow{-2}{*}{credit scoring  }  &$1,000$ &   \multirow{-2}{*}{credit records }   &  \multirow{-2}{*}{table  } &\newcite{misc_statlog_(german_credit_data)_144} &  \multirow{-2}{*}{Accuracy} \\

\midrule

CS & \multirow{3}{*}{firm disclosure }  &$240$ & \multirow{3}{*}{SEC filings}   &  \multirow{3}{*}{text}&\newcite{caoetal2023} & \multirow{3}{*}{Accuracy}  \\

FSR & &$3,931$ &  &  &\newcite{caoetal2020} &  \\

ITR &  &$1,196$ &  &  &\textbf{Our paper} & \\
\midrule


FinTerms-MCQ &\multirow{3}{*}{hallucination analysis}& $1,129$ & financial terms,Wikipedia & text & \textbf{Our Paper} & Accuracy \\
FinanceBench & & $150$ & financial documents & text,tables & \newcite{islam2023financebench} & \multirow{2}{*}{Human Evaluation} \\
FinTerms-Gen & & $150$ & financial terms,Wikipedia & text & \textbf{Our Paper} &  \\

\midrule

ConvFinQA &  &$3,892$ & &  &\newcite{chen-etal-2022-convfinqa} & \\
FinQA & \multirow{-2}{*}{numerical understanding }  &$8,281$ & \multirow{-2}{*}{earnings reports   } &   \multirow{-2}{*}{text,  table  }  &\newcite{chen-etal-2021-finqa} &  \multirow{-2}{*}{Exact Match} \\

\midrule
Finer-Ord &  &$1,080$ &news articles &  &\newcite{shah2023finer} &  \\
FiNER &\multirow{-2}{*}{named entity recognition  }  &$13,660$ &financial agreements &  \multirow{-2}{*}{text}   &\newcite{salinas-alvarado-etal-2015-domain} & \multirow{-2}{*}{Entity-F1 } \\

\midrule
ACL18 & \multirow{3}{*}{stock movement prediction }   &$27,053$ &  \multirow{3}{*}{tweets, historical prices }  & \multirow{3}{*}{text, time series }    &\newcite{xu2018stock} & \multirow{3}{*}{Accuracy}\\
BigData22 &   &$7,164$ &  & &\newcite{soun2022accurate} & \\

CIKM18 &  &$4,967$ &  &  &\newcite{wu2018hybrid} & \\

\midrule

FiQA-SA & \multirow{4}{*}{sentiment analysis}   &$11,730$ &news headlines, tweets &\multirow{4}{*}{text}    &\newcite{inproceedings} &  \multirow{4}{*}{Accuracy} \\
FOMC &  &$496$ &FOMC hawkish-dovish & &\newcite{shah-etal-2023-trillion} & \\
FPB &  &$48,450$ &news &  &\newcite{malo2013good} &   \\
Headline &  &$11,412$ &news headlines & &\newcite{sinha2020impact} & \\
\midrule

ECTSUM & \multirow{3}{*}{text summarization}   &$495$ &earning call transcript & \multirow{3}{*}{text}  &\newcite{mukherjee-etal-2022-ectsum} &\multirow{3}{*}{Rouge-score }  \\
EDTSUM &  &$2,000$ &news articles & &\newcite{zhou-etal-2021-trade} &  \\
Risk Eval & &$3,000$ &SEC articles & &\newcite{loukas-etal-2021-edgar} &  \\


\bottomrule
\end{tabular}}
\caption{The details of the downstream data. FinTerms-Gen is extracted from \newcite{Investopedia} and FinTerms-MCQ is generated using code from \newcite{ghosh-EtAl:2022:FNP}
}
\label{tab:downstream-data}
\end{table*}

A diverse array of downstream datasets is crucial for effective LLM performance benchmarking. In this work, we develop an extensive benchmark using existing and new datasets to evaluate our models. Our benchmark covers the following tasks: (1) chart understanding (CU), (2) sentiment analysis (SA), (3) named entity recognition (NER), (4) number understanding (NU), (5) text summarization (TS), (6) stock movement prediction (SMP), (7) credit scoring (CS), (8) firm disclosure (FD), and (9) hallucination analysis (HI).  Table~\ref{tab:downstream-data} summarizes all the datasets used in our evaluation, each along with the corresponding evaluation metric employed. We also provide more details about the datasets in Appendix \ref{dds}. 
\begin{figure}[t]
\includegraphics[width=\columnwidth]{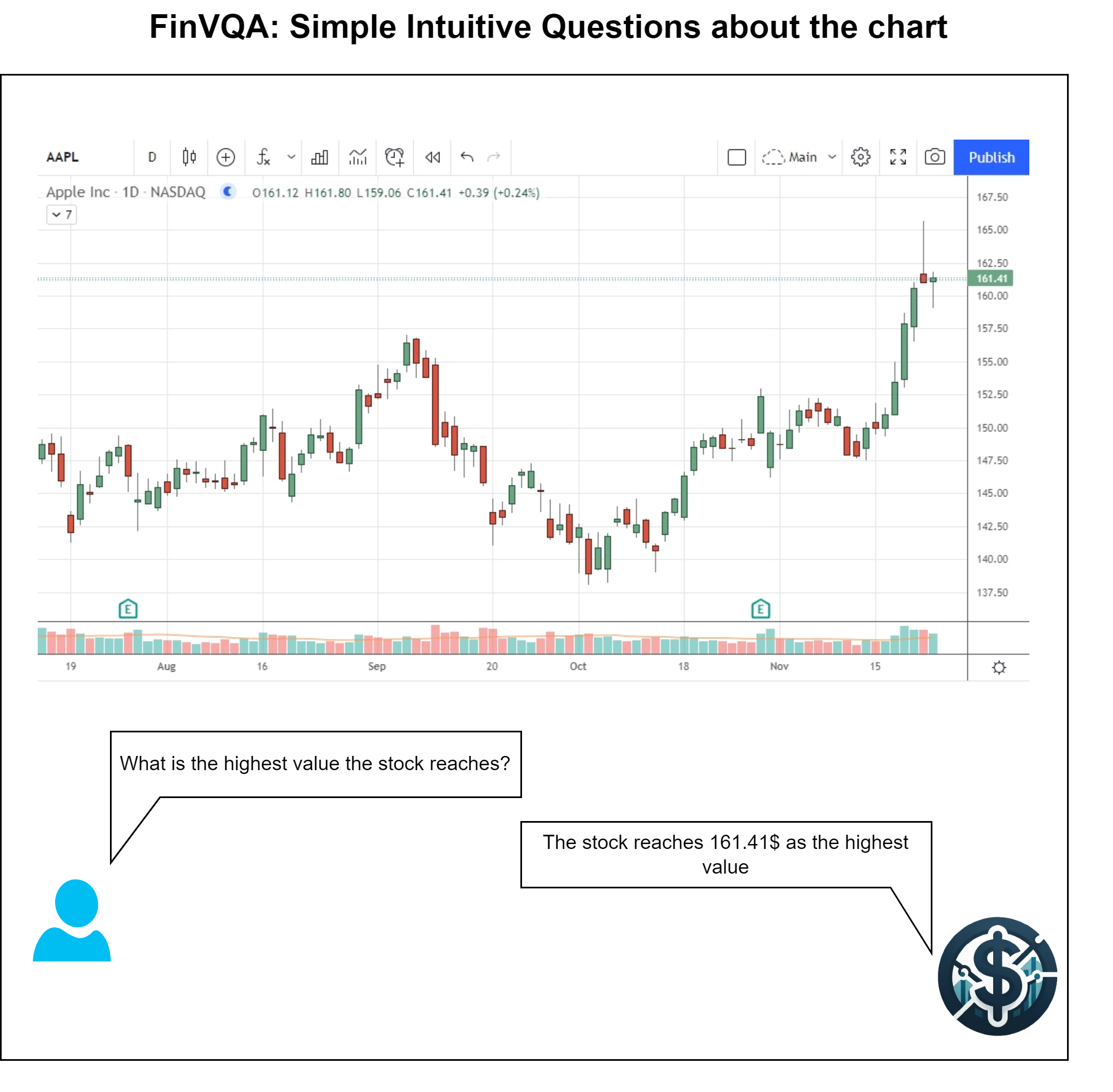}
\centering
\caption{Illustration of the FinVQA Dataset to understand model's proficiency in Chart Understanding Tasks. The figure demonstrates the model's ability to interpret stock price movements and identify peak values from a candlestick chart.} \label{fig:vl-fin}
\end{figure}

\section{Fintral} \label{models}

We use \textit{Mistral-7B-v0.1}~\cite{jiang2023mistral} as our base model for further development, due to its strong performance and employment of a BPE tokenizer that segments numbers into single digits, which is suitable for numerical tasks. 

\noindent\textbf{Domain-Specific Pretraining}
We further pretrain Mistral-7B-v0.1 on our 20 billion token FinSet financial data described in Section~\ref{data}.  
We perform pretraining with 
flash attention 2~\cite{dao2023flashattention2}. 
We employ a sequence length of up to 8k tokens, thus accommodating long financial documents. We use LoRA~\cite{hu2021lora} for pretraining and train the model for one
epoch with a learning rate of $2.5e^{-5}$. Pretraining takes 80 hours on four 40GB A100 GPUs. 

\begin{figure}[htb]
	\begin{center}
\begin{tikzpicture}[node distance=12mm]
	\node (n1) [prompt] {You are a \texttt{\clp{financial expert}} specializing in the nuanced analysis of financial statements and a wide array of data-driven financial tasks. For each prompt you are given, think step by step. Sometimes, you must extract relevant information to proceed with the problem. \\
        \textbf{Instructions:} 
        \clr{\texttt{If any options are specified, ensure that your answer is one of the options specified. \\}}
		\phantom{new} \clr{\texttt{Do not explain why you think the answer is correct. \\} }
  
	\textbf{Context:} \clb{\texttt{text} + \texttt{table} + \texttt{image}} \\
        \textbf{Retrieval:} \clb{\texttt{Retrieved relevant information}} \\
	\textbf{Question:} \clb{\texttt{Task based question?}} \\
        \textbf{Answer:}
}; 

\node (n3) [response, below=of n1]
	      {\texttt{Answer.}};
\draw [arrow] (n1) --  (n3);
\node[above left=0.25cm and -3.25cm of n3] (image) {\includegraphics[height=0.9cm]{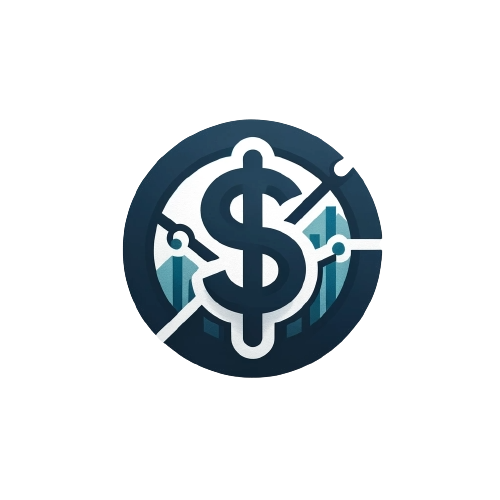}};
\node[below= -0.25cm of image] {\scriptsize FinTral};
	\node [fill=black, minimum width=0.05cm, minimum height=0.05cm, node distance=3mm, below left=0.25cm and -0.4cm of n3](c1) {};
	\node [right= -0.05cm of c1](cn1) {\scriptsize Signifier};
	\node [fill=cyan!05!purple, minimum width=0.05cm, minimum height=0.05cm, right= -0.01cm of cn1](c2) {};
	\node [right= -0.05cm of c2](cn2) {\scriptsize \clp{Memetic proxy}};
	\node [fill=cyan!05!red, minimum width=0.05cm, minimum height=0.05cm, right= -0.01cm of cn2](c3) {};
	\node [right= -0.05cm of c3](cn3) {\scriptsize \clr{Constraining behavior}};
	\node [fill=blue, minimum width=0.05cm, minimum height=0.05cm, right= -0.01cm of cn3](c4) {};
	\node [right= -0.05cm of c4](cn4) {\scriptsize \clb{Input}};
\end{tikzpicture}
	\end{center}
\caption{FinTral prompting method}
\label{fig:FinBEAT_prompt}
\end{figure}
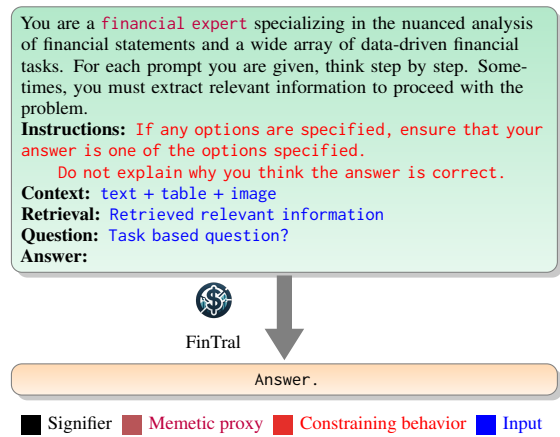

\noindent\textbf{Prompting for Financial LLMs} \label{promptings}
We employ a prompting method suited for a financial LLM with multimodal capabilities. The model is assigned a memetic proxy~\cite{reynolds2021prompt} as a financial expert signifying key expected behaviors, encouraged to think step by step, and that consider diverse inputs which may be texts, tables, or images. This is followed by a strategic retrieval of pertinent information, ensuring the model's focus aligns with the query's requirements. The model then engages with a task-based question, demanding an application of the model's financial expertise and analytical thinking. This structured approach is pivotal in eliciting focused answers from the model, especially in complex financial scenarios. The application of constraints further refines the model's output, leading to enhanced accuracy and context-appropriate responses. A visual representation of~\ourmodel~'s prompting method is depicted in Figure~\ref{fig:FinBEAT_prompt}. 

\noindent\textbf{Instruction Tuning}
We use our instruction tuning dataset described in section~\ref{instruction_data} to perform instruction finetuning on our pretrained model.\footnote{We standardize all the datasets to have the same format of prompting, as explained earlier.
} 
We adopt QLoRA
to perform instruction finetuning using all the linear layers as target modules as this gives us a performance that is close to full fine-tuning \cite{dettmers2023qlora}.

\noindent\textbf{Alignment with AI Feedback}
Large language models may fail to respond well to natural prompts even after instruction fine-tuning. 
 To address this challenge, we use direct preference optimization (DPO) \cite{rafailov2023direct} which allows us to preferentially tune the model without the usage of a reward model. \citet{tunstall2023zephyr} introduces a method to use LoRA to train LLMs using DPO objective. This is known as distilled direct preference optimization (dDPO).
\footnote{We use the scripts provided by ~\citet{tunstall2023zephyr} to train our dDPO model.} We describe how we generate the binarized preference data for this process in Section~\ref{feedback_data}.


\noindent\textbf{Multimodal Instruction Tuning}
Once we teach our model to handle various financial queries, we also empower it with visual understanding. This is done using the architecture suggested by~\citet{liu2023improved}.
Specifically, we add an \texttt{<image>} token to our prompt and replace the \texttt{<image>} token with its image embedding after tokenization.
We use a CLIP model~\cite{radford2021learning} as our vision encoder and a 2-layer MLP visual abstractor, allowing us to convert image inputs into text embeddings fed to the LLM. 

\noindent\textbf{Tool Usage} \label{tools}
In addressing the inherent challenges faced by LLMs
in dealing with quantitative tasks, we integrate tools~\cite{schick2023toolformer} to our model. These tools enable the LLM to offload mathematically intensive tasks to a more suitable computational environment. For instance, functions such as \texttt{Add()}, \texttt{Subtract()}, and \texttt{Multiply()} are used by model to generate outputs in a structured format interpretable as Python function calls, thereby enhancing model accuracy in financial applications.

\noindent\textbf{Retrieval Augmented Generation (RAG)}
As shown in~\citet{zhang2023enhancing} for financial sentiment analysis, using retrieval augmented generation (RAG) can significantly boost performance. To better facilitate our tool usage and, in some cases, text extraction from complex data, we deploy a RAG system employing the BGE~\cite{bge_embedding} models, which are SoTA for document retrieval. This is useful for LLMs since users commonly ask out-of-domain questions. We use 30,000 financial documents derived from multiple sources covering January 1, 2022 to September 30, 2023. We use the chain of retrieval, as shown in Figures ~\ref{fig:chain_of_ret} and its example is provided in Figure~\ref{fig:FinBEAT_prompt_example}. 




\section{Experiments} \label{results}
\begin{table*}[ht]\centering
\scriptsize
\resizebox{\textwidth}{!}{%
\begin{tabular}{lccccccccc}\toprule
\textbf{Model} &\textbf{Type} &\textbf{SA} &\textbf{NER} &\textbf{NU} &\textbf{TS} &\textbf{SMP} &\textbf{CS} &\textbf{FD} &\textbf{Average} \\\midrule
FinMA-7B-trade & $\spadesuit$ &0.20 &0.00 &0.00 &0.08 &0.46 &0.39 &0.00 &0.16 \\
Llama-2-7b-hf & $\clubsuit$ &0.26 &0.00 &0.00 &0.00 &0.48 &0.50 &0.09 &0.19 \\
Mistral-7B-v0.1 & $\clubsuit$ &0.25 &0.00 &0.00 &0.05 &0.49 &0.52 &0.09 &0.20 \\
Vicuna-7B & $\diamondsuit$ &0.54 &0.01 &0.00 &0.20 &0.46 &0.39 &0.00 &0.23 \\
Mistral-7B-Instruct-v0.1 & $\diamondsuit$ &0.49 &0.00 &0.00 &0.30 &0.49 &0.48 &0.29 &0.29 \\
Llama-2-13b-chat-hf & $\heartsuit$ &0.58 &0.02 &0.00 &0.30 &0.50 &0.52 &0.31 &0.32 \\
FinMA-7B & $\spadesuit$  &0.72 &0.38 &0.16 &0.29 &0.46 &0.29 &0.00 &0.33 \\
Llama-2-7b-chat-hf & $\heartsuit$ &0.54 &0.07 &0.00 &0.31 &0.52 &0.56 &0.32 &0.33 \\
FinMA-7B-full & $\spadesuit$ &0.78 &0.35 &0.12 &0.35 &0.51 &0.29 &0.30 &0.38 \\
\textbf{FinTral-INST} & $\diamondsuit$ &\underline{0.81} &0.40 &0.02 &0.40 &\underline{0.53} &0.61 &0.66 &0.49 \\
ChatGPT (gpt-3.5-turbo) &$\heartsuit$ &0.70 &0.53 &\underline{0.58} &0.59 &0.53 &0.31 &0.52 &0.53 \\
\textbf{FinTral-DPO} &$\heartsuit$ &\textbf{0.82} &\underline{0.70} &0.15 &\underline{0.60} &\textbf{0.54} &\underline{0.62} &\underline{0.67} &\underline{0.59} \\
GPT-4 (gpt-4-0613) & $\heartsuit$ &0.79 &\textbf{0.80} &\textbf{0.63} &\textbf{0.65} &\textbf{0.54} &\textbf{0.70} &\textbf{0.73} &\textbf{0.69} \\
\bottomrule
\end{tabular}
}
\caption{Comparative analysis of LLMs on diverse tasks. Models in bold are introduced in this paper.  This analysis includes \textbf{SA:} Sentiment Analysis, \textbf{NER:} Named Entity Recognition, \textbf{NU:} Number Understanding, \textbf{TS:} Text Summarization, \textbf{SMP:} Stock Movement Prediction, \textbf{CS:} Credit Scoring, and \textbf{FD:} Firm Disclosure. 
}
\label{tab: results}
\end{table*}
We conducted multiple experiments to illustrate the efficacy of the methods described in section \ref{models}. We evaluated our model on the downstream tasks described in section \ref{donwstream_tasks_data}. 
Symbols in the following tables indicate the types of models: $\clubsuit$, $\spadesuit$, $\diamondsuit$, $\heartsuit$, $\bigstar$ and, $\blacksquare$ represent the pre-trained model, the fine-tuned model, the instruction fine-tuned model, the RL-Tuned Models,  tools, and retrieval, respectively. We then performed a hallucination index accuracy check to assess how well our model mitigates one of the biggest challenges for LLMs. 

We introduce three versions of our model. Firstly, FinTral-INST is our instruction-fine-tuned model obtained by fine-tuning our pre-trained model. Note that we do not assess the performance of the pre-trained model as it serves as an intermediate step to the instruction fine-tuning model. Secondly, We introduce FinTral-DPO, which has been further trained based on FinTral-INST utilizing reinforcement learning using AI feedback with the dDPO objective. Then, we introduce our FinTral-DPO-T\&R, which combines the FinTral-DPO with tools and retrieval. 

We also compare performance of our models to nine other baselines LLMs. These are LLama-2 \cite{touvron2023llama}, Mistral \cite{jiang2023mistral}, three versions of FinMA \cite{xie2023pixiu}, Vicuna \cite{vicuna2023}, ChatGPT \cite{openai2023chatgpt}, GPT-4 \cite{gpt4}.

\subsection{Instruction Tuning and Model Alignment}

As seen from Table~\ref{tab: results}, our instruction fine-tuned model FinTral-INST outperforms all pretrained and fine-tuned open-source models with an average score of 0.49.
One of the causes of concern here is the tasks that require a specific format as the output, like the numerical understanding and NER tasks. 
We see that in some instances, the model struggles to follow instructions and often deviates from what the task asks for. 

Furthermore, models that have undergone reinforcement learning with AI feedback (RLAIF), like FinTral-DPO, ChatGPT, and GPT-4, show even more marked improvements. Adding RLAIF dramatically boosts performance to the average score of 0.59, resulting in FinTral-DPO outperforming ChatGPT.  

GPT-4, in particular, stands out with the highest average score, indicating its robust performance across a diverse set of tasks. Its high NER, NU, and FD scores suggest exceptional capabilities in understanding complex text, identifying specific entities, and interpreting numerical data. 

\subsection{Retrieval and Tools Usage}

As detailed in section \ref{models}, the use of retrieval and tools plays a pivotal role in enhancing the capabilities of our model, FinTral-DPO-T\&R, similar to their impact on GPT-4. Integrating these features into these models allows the models to access a broader range of information and apply more specialized processing techniques, leading to significant improvements in performance across various tasks.
In the case of FinTral-DPO-T\&R, combining the FinTral-DPO model with retrieval and tool capabilities has proven particularly effective. The FinTral-DPO model's ability to follow instruction prompts accurately enables seamless integration with external tools and retrieval data. 
The performance of GPT-4-Turbo, with its latest update incorporating tools and retrieval, is also noteworthy. 

In 5 downstream tasks, FinTral-DPO-T\&R outperformed GPT-4, while GPT-4 surpassed FinTral-DPO-T\&R in two downstream tasks. Since GPT-4 has done exceptionally well in those two tasks, its average performance is slightly better than FinTral-DPO-T\&R (0.72 vs. 0.70, as shown in table \ref{tab:tr results}).
The edge that FinTral-DPO-T\&R and GPT-4 have over other models is a testament to the potential of combining sophisticated AI models with additional data and tool integration for more refined and accurate outputs.
\begin{table*}[!htp]\centering
\scriptsize
\resizebox{\textwidth}{!}{%
\begin{tabular}{lccccccccc}\toprule
\textbf{Model} &\textbf{Type} &\textbf{SA} &\textbf{NER} &\textbf{NU} &\textbf{TS} &\textbf{SMP} &\textbf{CS} &\textbf{FD} &\textbf{Average} \\\midrule
Mistral-7B-Instruct-v0.1 & $\diamondsuit$ &0.49 &0.00 &0.00 &0.30 &0.49 &0.48 &0.29 &0.29 \\
Llama-2-7b-chat-hf &$\heartsuit$ + $\bigstar$ + $\blacksquare$ &0.54 &0.07 &0.00 &0.31 &0.52 &0.56 &0.32 &0.33 \\
\textbf{FinTral-INST} &$\diamondsuit$ &0.81 &0.40 &0.02 &0.40 &0.53 &0.61 &0.66 &0.49 \\
ChatGPT (gpt-3.5-turbo-1106) & $\heartsuit$ &0.70 &0.53 &0.58 &0.59 &0.53 &0.31 &0.52 &0.53 \\
\textbf{FinTral-DPO} &$\heartsuit$ &\underline{0.82} &0.70 &0.15 &0.60 &0.54 &0.62 &0.67 &0.59 \\
\textbf{FinTral-DPO-T\&R} &$\heartsuit$ + $\bigstar$ + $\blacksquare$ &\textbf{0.83} &\textbf{0.83} &\underline{0.60} &\textbf{0.72} &\textbf{0.56} &\underline{0.62} &\textbf{0.75} &\underline{0.70} \\
GPT-4-Turbo (gpt-4-1106-preview) & $\heartsuit$ + $\bigstar$ + $\blacksquare$ &0.79 &\underline{0.80} &\textbf{0.83} &\underline{0.65} &\underline{0.54} &\textbf{0.70} &\underline{0.73} &\textbf{0.72} \\
\bottomrule
\end{tabular}}
\caption{Comparative analysis of LLMs using external tools on diverse tasks. 
}
\label{tab:tr results}
\end{table*}
 
\subsection{Multimodal Evaluation}



To evaluate our financial multimodal model, we use ChartQA and our FinVis datasets. We compare various state-of-the-art multimodal large language models (MLLMs) such as GPT-4V \cite{openai2023gpt4}, Gemini-Pro \cite{geminiteam2023gemini}, Qwen-VL-Plus \cite{bai2023qwenvl}, LLaVa-NEXT \cite{liu2024llavanext}, and our FinTral-VL model which comprises of CLIP and FinTral-DPO. As Table~\ref{tab: vl} shows, GPT-4V performs best, with scores of 0.79 in ChartQA and 0.89 in FinVis, averaging 0.84. Gemini-Pro follows closely, with a consistent performance across both datasets, scoring an average of 0.78. Other models like Qwen-VL-Plus, FinTral-VL, and LLaVa-NEXT show varying degrees of efficacy: Qwen-VL-Plus performing notably better in ChartQA (0.78) than in FinVQA (0.64), while FinTral-VL and LLaVa-NEXT trail behind, indicating areas for potential improvement in their visual data interpretation capabilities. FinTral-VL performs well on the FinVQA dataset, making it highly suited for multimodal financial usage. Figure \ref{fig:vl-fin-fig} shows examples of models' outputs on questions from the FinVQA dataset. 

\begin{table}[!htp]\centering
\scriptsize
\resizebox{\columnwidth}{!}{%
\begin{tabular}{llccc}\toprule
\textbf{Method} &\textbf{LLM} &\textbf{ChartQA } &\textbf{FinVQA} &\textbf{CU} \\\toprule
\multicolumn{5}{c}{\textit{\textbf{Closed-source API}}} \\
\midrule
Gemini-Pro &{~~~~~~~-} &0.74 &0.82 &0.78 \\
QwenVL-Plus &{~~~~~~~-} &0.78 &0.64 &0.71 \\
GPT-4V &{~~~~~~~-} &\textbf{0.79} &\textbf{0.89}&\textbf{0.84} \\
\midrule
\multicolumn{5}{c}{\textit{\textbf{Open-source MLLMs}}} \\ \midrule
LLaVA &Vicuna-7B &0.12 &0.25 &0.19 \\
InstructBLIP &Vicuna-7B &0.34 &0.23 &0.29 \\
LLaVA-1.5 &Vicuna-13B &0.44 &0.32 &0.38 \\
Qwen-VL-Chat &Qwen-7B &0.53 &0.34 &0.44 \\
LLaVa-NEXT &Yi-34B &\textbf{0.65} &0.58 &0.62 \\
FinTral-VL (ours) &FinTral-DPO &0.63 &\textbf{0.75} &\textbf{0.69}\\
\bottomrule
\end{tabular}}
\caption{Comparison with available MLLMs on Chart Understanding datasets.}
\label{tab: vl}
\end{table} 
\subsection{Financial Hallucination Evaluation} \label{error}


Since financial hallucinations can be complex to measure, we have used three different methods and datasets to quantify hallucinations. We first assess how much models hallucinate in selecting definitions of financial terms. We then conduct human evaluations of the appropriateness of responses from top LLM models based on our first task. Finally, we       
evaluated them on the Finance Bench \cite{islam2023financebench} dataset, a complex numerical question-answering dataset requiring mathematical tools and retrieval. 

\noindent\textbf{FinTerms-MCQ} 
In FinTerms-MCQ dataset, we convert definitions of financial terms from \citet{Investopedia} to a multiple choice format using the right definition and three other closely related definitions. We then ask the models to select the right definition. We derive a hallucinations index (HI), defined as the proportion of correctly generated definitions by each model (higher is better), based on the models' performance in this MCQ task. As seen in Table \ref{tab: model_halo}, the models' performances on the HI vary significantly. 
GPT-4 and ChatGPT lead the pack with exceptionally high scores of 98\% and 95\%, respectively. All three of our models perform better than the other open-source LLMs. In particular, FinTral-DPO-T\&R show a strong performance with an HI of 97\%.

\begin{figure}[!htp]
\centering
\includegraphics[width=1\columnwidth]{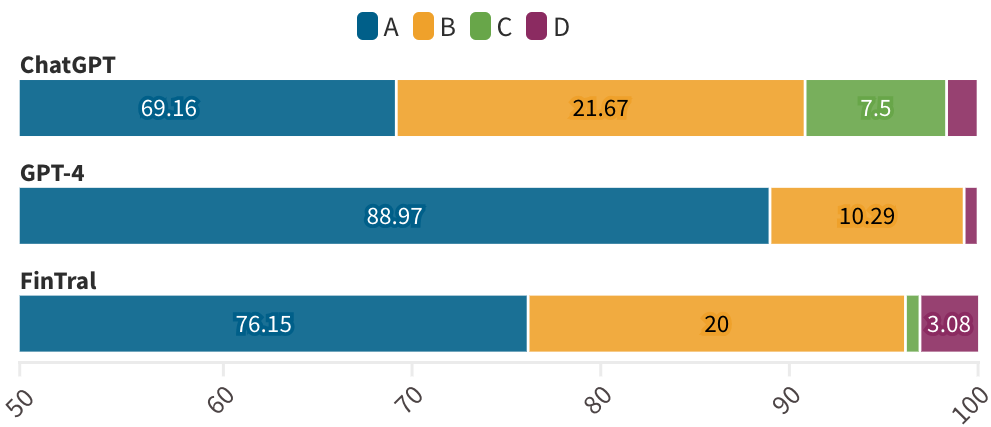}
\caption{Human Evaluation on FinTerms Dataset. \textbf{\textit{FinTral:}} is our FinTral-DPO-T\&R. Each bar is segmented into four colors representing the quality of responses: \colorSquare{colorA} \textbf{A}: correct and satisfying response \colorSquare{colorB} \textbf{B}: acceptable response with minor imperfection, \colorSquare{colorC} \textbf{C}: responds to the instruction but has significant errors, \colorSquare{colorD} \textbf{D}: irrelevant or invalid response. 
} \label{fig:human_eval}
\end{figure}

\begin{table}[!ht]\centering
\scriptsize
\resizebox{\columnwidth}{!}{%
\begin{tabular}{lcc}\toprule
\textbf{Model} &\textbf{Type} &\textbf{HI} \\\midrule
FinMA-7B-trade &$\spadesuit$ &$0.28$ \\
Vicuna-7B &$\diamondsuit$ &$0.55$ \\
Llama-2-7b &$\clubsuit$ &$0.64$ \\
FinMA-7B &$\spadesuit$ &$0.64$ \\
Mistral-7B &$\clubsuit$ &$0.67$ \\
Llama-2-7b-chat &$\heartsuit$ &$0.70$ \\
Llama-2-13b-chat &$\heartsuit$ &$0.75$ \\
Mistral-7B-Instruct &$\diamondsuit$ &$0.76$ \\
FinMA-7B-full &$\spadesuit$ &$0.80$ \\
\textbf{FinTral-INST} &$\diamondsuit$ &$0.82$ \\
\textbf{FinTral-DPO} &$\heartsuit$ &$0.88$ \\
ChatGPT &$\heartsuit$ &$0.95$ \\
\textbf{FinTral-DPO-T\&R} &$\heartsuit$ + $\blacksquare$ &\ul{$0.97$} \\
GPT-4-Turbo &$\heartsuit$ + $\blacksquare$ &\textbf{0.98} \\
\bottomrule
\end{tabular}}
\caption{Comparison of various models based on Hallucinations Index (HI). This index represents the proportion of correctly generated definitions by each model (higher is better). 
}
\label{tab: model_halo}
\end{table}
\noindent\textbf{FinTerms-Gen} 
In Table~\ref{tab:halo_eg}, we show an example of how popular LLMs, like ChatGPT, hallucinate in the financial domain. We generate answers to questions related to the financial terms in the FinTerms-Gen dataset (n=150, see Table~\ref{tab:downstream-data}) using the three models with best performance on FinTerms-MCQ (i.e, GPT-4, ChatGPT, and FinTral-DPO+T\&R). We then ask two humans, each with at least four years of background in finance, to label the responses with one of the four correctness tags shown in Figure~\ref{fig:human_eval}. The two annotators agree with a Cohen’s kappa (\textit{K}) of $0.85$. As Figure \ref{fig:human_eval}\footnote{We use only Q\&A pairs where both annotators agree (n=128 pairs) for this analysis.} shows our~\ourmodel~-DPO-T\&R produces more correct and satisfying responses (category A in Figure~\ref{fig:human_eval}) than ChatGPT but falls short of GPT-4.

\begin{figure}[!ht]
\centering
\includegraphics[width=1\columnwidth]{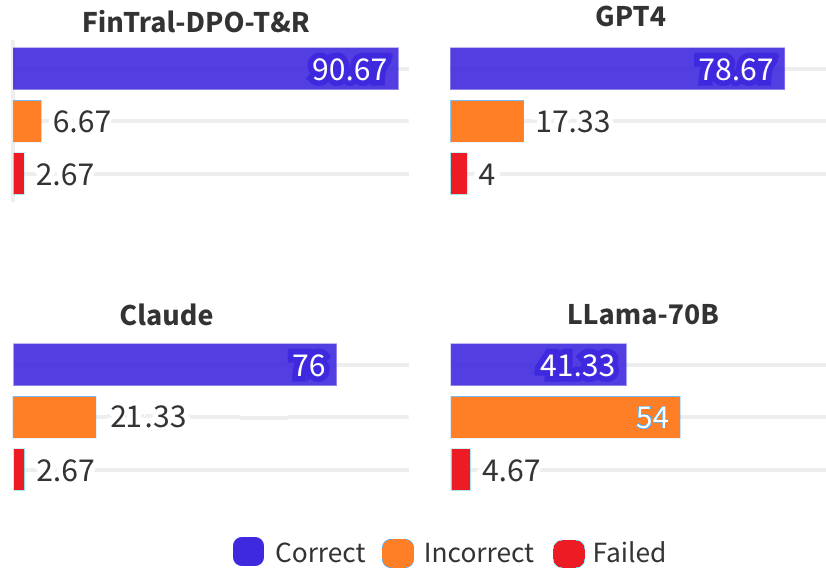}

\caption{Performance comparison of various models on the FinanceBench dataset. Each model's percentage of correct, incorrect, and failed responses is shown. FinTral-DPO-T\&R and GPT4 outperform other models, with LLama-70B having the highest failure rate. } \label{fig:finacebench}
\end{figure}

\noindent\textbf{Finance Bench} Finance Bench \cite{islam2023financebench} is a proprietary dataset designed to assess the capabilities of LLMs in the context of open-book financial question answering (QA). While the full version includes 10,231 questions related to publicly traded companies, each accompanied by evidence strings and relevant answers, we evaluate our models using FinanceBench's open-source sample of 150 questions as provided in \citet{islam2023financebench} using the same methodology adopted by the authors. As presented in Figure \ref{fig:finacebench}, the FinTral-DPO-T\&R performs very well on this dataset, outperforming the other models, GPT-4 \cite{gpt4}, Claude \cite{claude-model-card}, and Llama-70B \cite{touvron2023llama}, evaluated in~\citet{islam2023financebench}. Using retrieval and tools in FinTral-DPO-T\&R proves its efficiency and puts the model ahead of all the other models. 

\section{Discussion} \label{discuss}




\noindent\textbf{Advancements in financial LLMs}
FinTral leverages extensive datasets and diverse training methods, including instruction fine-tuning and RLAIF, to enhance its analysis of complex financial data across multiple modalities. The integration of advanced tools further augment its financial capabilities.

\noindent\textbf{Reducing model hallucinations}
FinTral combats financial hallucinations by pretraining with up-to-date, clean financial data and employing RLAIF and retrieval methods, enhancing model accuracy and reliability.

\noindent\textbf{Human-AI collaboration in financial decision-making}
Enhancing FinTral's real-time adaptability to financial markets through dynamic data retrieval and live data analysis can significantly boost its predictive accuracy and assist in informed decision-making. Figure \ref{fig:applications} shows how this model can be used in real world. 

\subsection{Model Shortcomings Analysis}

Each iteration of FinTral was designed to progressively address the shortcomings of its predecessors, focusing on enhancing financial domain-specific knowledge and instruction compliance while minimizing the occurrence of hallucinations.

The initial versions, starting from Mistral-7B and the subsequent FinTral, showed improved domain knowledge but struggled with maintaining instruction accuracy and often produced misleading outputs. FinTral-INST, despite advancements, continued to face challenges in adhering to specific task formats and response coherence. This was partly mitigated in FinTral-DPO, which better adhered to the instructed formats by introducing direct preference optimization.

The most advanced iteration, FinTral-DPO-T\&R, integrates tools and retrieval capabilities to further refine performance, especially in complex financial tasks requiring extensive data integration and computational power. This version demonstrates substantial improvements in handling detailed financial analyses and proves highly competitive, even with leading models like GPT-4, particularly in domains requiring rigorous financial insight.

\section{Conclusion} \label{conclusion}

We presented~\ourmodel, an advanced multimodal financial language model with remarkable capabilities. Key advancements include integrating textual, numerical, and visual data, a training pipeline with various finetuning capabilities, and employment of tools and retrieval mechanisms. The model effectively addresses challenges like financial hallucination, evidenced by high performance in various financial tasks compared to baseline models. 
The achievements of FinTral hold a great potential for financial models of a moderate size (e.g., 7B).

\section{Limitations}\label{sec:limit}

While FinTral represents a significant advancement in the realm of financial large language models (LLMs), it is important to acknowledge inherent limitations:

\begin{enumerate}
    \item \textbf{Domain-Specific Adaptability:} Tailored for the financial domain, FinTral may not perform as effectively outside its trained scope, potentially limiting its generalizability.
    
    \item \textbf{Handling of Real-Time Data:} While designed for real-time analysis, the model's predictive accuracy depends on the timeliness and accuracy of incoming data, which may be affected by rapidly changing market conditions.
    
    \item \textbf{Maintenance and Updating:} Continuous updating and maintenance are required to keep the model relevant and effective in evolving financial markets and regulations.
\end{enumerate}

Acknowledging these limitations is crucial for the responsible deployment and continued development of FinTral and similar financial LLMs.

\section{Ethics Statement}\label{sec:ethics}

\noindent\textbf{Energy Efficiency.} Our FinTral models, similar to many large language models (LLMs), required significant training time and computational resources, and thus are not particularly energy efficient. We acknowledge this as a critical issue and advocate for ongoing research into developing more energy-efficient models. 

\noindent\textbf{Data.} Our pretraining datasets are collected from public domains, encompassing a wide range of financial topics and sources. While these datasets provide comprehensive coverage for financial language modelling, we must be aware of the potential biases and limitations inherent in publicly available data, ensuring our model remains as objective and unbiased as possible.

\noindent\textbf{Data Copyright.} We emphasize that all datasets used, including those from SEC filings, news sources, and social media, are collected from publicly available sources. We confirm that our data collection process respects the copyrights of these sources and does not infringe upon any proprietary data.

\noindent\textbf{Model Release.} We are considering releasing our models and evaluation data (FinSET) responsibly. Given the sensitive nature of financial data and the potential for misuse, we will implement strict guidelines and conditions for the use of FinTral, particularly in real-world applications. This includes clear guidelines on ethical usage and the avoidance of deployment in contexts that could lead to unethical practices such as market manipulation or privacy violations.

\noindent\textbf{Privacy.} FinTral is developed using publicly available data, which mitigates concerns regarding personal information leakage. However, given the sensitive nature of financial data, we have taken extra precautions to ensure that no identifiable personal or corporate financial information is retrievable from our trained models.

\noindent\textbf{Human Annotation.} The human annotators involved in this project are professionals with expertise in finance and natural language processing. No sensitive or personally identifiable data was used in the annotation process, adhering to ethical guidelines and data privacy standards. The human annotators are co authors on this paper. 

\noindent\textbf{Bias Analysis.} We recognize that any language model can inadvertently perpetuate biases present in its training data. In FinTral's case, potential biases might be related to financial markets, regions, or corporate entities. We conducted thorough analysis to identify and mitigate such biases, ensuring that our model's outputs are as fair and unbiased as possible. However, users should remain aware of these potential biases, especially when applying the model to real-world scenarios.

\noindent\textbf{Applications.} While FinTral offers advanced capabilities for financial analysis, like any powerful tool, it can be misused. It's crucial to emphasize responsible usage, particularly in sensitive financial contexts. Users should avoid deploying FinTral for speculative trading, market manipulation, or any activity that could contravene financial regulations or ethical standards. Conversely, FinTral has the potential for beneficial applications such as financial education, research, and improving the accessibility of financial information.

\noindent\textbf{AI usage.} It's pertinent to acknowledge the role of AI tools such as ChatGPT in our project. Specifically, ChatGPT was utilized minimally and primarily for grammar corrections in our documents. This use was strictly confined to enhancing linguistic accuracy and improving the readability of our written materials. It's important to clarify that the core research, analysis, and development were conducted independently by our team.

\bibliography{custom}
\normalem
\newpage
\appendix

\clearpage
\appendixpage
\addappheadtotoc
\numberwithin{figure}{section}
\numberwithin{table}{section}


\section{Detailed Related Works} \label{subsec:literature}
\noindent\textbf{Financial NLP Models and Their Challenges}
There have been successful applications of traditional Natural Language Processing (NLP) techniques a range of finance related problems. These include named entity recognition \cite{salinas-alvarado-etal-2015-domain} 
sentiment analysis of financial news \cite{souma2019enhanced,araci2019finbert}, event extraction \cite{yang2018dcfee,zheng2019doc2edag}, generating financial reports \cite{chapman2022towards}, and text summarization in a financial context \cite{la2020end}.

However, deploying NLP models for domain-specific tasks in the financial sector faces several distinct challenges. Firstly, the complex and jargon-rich nature of financial language poses a significant barrier in achieving desirable performance from the models, often leading to a gap in understanding the domain-specific documents \cite{mik2017smart}. Secondly, the scarcity of annotated datasets, combined with the high costs associated with data annotation, in finance, hinders the advancement of these models. Thirdly, existing NLP models often fall short in inferential capabilities, particularly in critical tasks such as risk assessment and making informed decision-making in investment contexts \cite{liu2023fingpt}. Additionally, the dynamic nature of financial markets requires models to be capable of real-time analysis, a feature that many current models do not possess. Numerical information processing, a common element in financial documents filled with figures and symbols, also poses a significant challenge for understanding financial documents. The challenge is further exacerbated by the fact that many graphs and figures are in image formats in these documents. Lastly, the wide-spread adaptability of many NLP models remains limited, as they are typically optimized for a particular single-task function and lack the ability to generalize across multiple tasks \cite{mishra2021cross}. In light of these challenges, it is imperative for ongoing research efforts to develop more advanced, versatile, and robust NLP models tailored to the dynamic and complex requirements for financial document undertanding.  

\noindent\textbf{Financial Large Language Models }
Finance has witnessed significant advancements in large language models, starting with the introduction of FinBERT \cite{araci2019finbert}. This early contribution sets a precedence for using pre-trained language models in financial sentiment analysis, demonstrating marked improvements in performance metrics. In 2023, a series of groundbreaking models have further propelled the field. BloombergGPT \cite{wu2023bloomberggpt} emerged as a 50-billion parameter model trained on an extensive financial data corpus. Its training on a diverse dataset enabled it to excel in financial tasks while maintaining robust performance in general LLM benchmarks. PIXIU \cite{xie2023pixiu} followed, presenting a comprehensive framework with a financial LLM fine-tuned with instruction data. PIXIU was a crucial development in advancing the open-source development of financial AI, combining a novel instruction dataset and an evaluation benchmark for financial LLMs. The same year saw the introduction of Instruct-FinGPT \cite{zhang2023instructfingpt}, which utilized instruction tuning to enhance financial sentiment analysis. This model particularly excelled in scenarios requiring deep numerical understanding and contextual comprehension. Another significant advancement was GPT-FinRE \cite{rajpoot2023gptfinre}, focusing on financial relation extraction using in-context learning. This model demonstrated high effectiveness and accuracy by employing two distinct retrieval strategies. Adding to the multimodal capabilities in financial LLMs, FinVis-GPT \cite{wang2023finvisgpt} was proposed, designed explicitly for financial chart analysis. This model leveraged the power of LLMs along with instruction tuning and multimodal capabilities, showcasing superior performance in related tasks. GPT-InvestAR \cite{gupta2023gptinvestar} aimed to enhance stock investment strategies by analyzing annual reports using LLMs. This approach yielded promising results in outperforming traditional market returns, highlighting the potential for LLMs in investment strategies. InvestLM \cite{yang2023investlm} showed strong capabilities in understanding economic text and providing practical investment advice.
 With retrieval-augmented LLMs \cite{zhang2023enhancing} addressed the challenges of applying LLMs directly to economic sentiment analysis, achieving considerable performance gains. FinGPT \cite{wang2023fingpt} focused on creating a benchmark for Instruction Tuning of LLMs in financial datasets, emphasizing the integration challenges and potential solutions for GPT-based models specialized in the financial domain. \citeauthor{sarmah2023reducing} \citeyearpar{sarmah2023reducing} reduced hallucination in information extraction from earning call transcripts and achieved improved the accuracy by combining retrieval-augmented generation techniques with metadata. FinLMEval \cite{guo2023chatgpt} assessed the performance of LLMs in financial natural language processing tasks, offering foundational evaluations for ongoing efforts to enhance LLMs in the financial domain. DISC-FinLLM \cite{chen2023discfinllm} introduced a Chinese financial LLM based on a Multiple Experts Fine-tuning Framework, showing improved performance in various monetary scenarios compared to baseline models. Lastly, the work on data-centric financial LLMs \cite{chu2023datacentric} presented a novel approach to better handle financial tasks with LLMs, emphasizing data pre-processing and pre-understanding, resulting in substantial performance improvements on economic analysis and interpretation tasks. These contributions collectively illustrate the rapid growth in utilization of LLMs and their tremendous potential in various financial applications, showcasing their capacities in revolutionizing financial analysis, forecasting, and decision-making processes. 

\section{Pretraining Data Details} \label{pretraindata}

\textbf{Common Crawl Data} \label{electra}
The Common Crawl dataset, specifically the C4 snapshot from 2019 to 2021, comprising over 10 billion files, was an initial broad data source. Text classification via the ELECTRA Finance domain-specific language model ensured that the dataset maintained a strong relevance to financial content. Rigorous domain filtering and data pruning were employed, isolating financial-specific texts and discarding irrelevant content. The final dataset consisted of 800 million documents, including 300 million English-only and 500 million multilingual files, providing a comprehensive base for financial analysis.

\noindent\textbf{News Scraping} Our approach extended to news scraping, particularly focusing on the period from July 2022 to July 2023. With 300 million data lines, this dataset allowed for in-depth analysis of market trends and financial narratives. The dataset encapsulated a global view of financial markets by integrating sources like Yahoo, Seeking Alpha, Eastmoney, and Yicai. This multi-source strategy ensured a robust, cross-referenced, and credible dataset. We used scrapers implemented in \cite{yang2023fingpt} to build out News datasets.

\noindent\textbf{SEC Filings} An exhaustive scrape of the EDGAR SEC database from 1993 to 2023 provided detailed records of accurate business, financial and accounting information from official filings. This dataset, exclusively in English, added substantial depth, allowing for analysis of historical market regulatory impacts and corporate financial maneuvers.

\noindent\textbf{Company Websites and Social Media} Further data were obtained from the top 5000 company websites and their social media presence on platforms like Facebook, Instagram, and Reddit. This dataset provided direct corporate communications and captured broader market sentiments and public perceptions, notably through an extensive scrape of the r/WallStreetBets Reddit community.

\section{Financial data cleaning and deduplication pipeline}\label{dedup} We started of gathering various text corpora shown in table \ref{tab: raw_data}, resulting in a dataset consisting of 2.9B documents. 
The data that we collected is not only unclean but also suffers from large-scale duplication. As shown by \cite{cerebras2023slimpajama}, using clean and deduplicated data is computationally efficient for model training. The data cleaning and deduplication pipeline for financial data begins with URL filtering, in which the raw data is initially processed. This crucial step ensures the inclusion of only pertinent URLs, enhancing the dataset's quality by excluding irrelevant or unsuitable sources. Once the URLs are streamlined, the Text extraction phase commences, whereby contents of documents from the selected URLs are meticulously extracted, filtering out images while maintaining the large dataset scale. Following this, the language identification phase excluded non-English documents by categorizing them based on the language of their tokens. Subsequently, the pipeline further refines the data through Document-wise domain-based filtering, narrowing down to 100 billion tokens pertinent to the financial domain by excluding 55B-token non-financial documents. Recognizing the importance of data privacy and relevance, the pipeline incorporates removing sensitive information, which is done using a classifier built using FinBERT \cite{araci2019finbert}. Line-wise corrections enhance accuracy and filter out 5B tokens of sensitive information. An extensive Fuzzy deduplication process reduces the data to 38 billion tokens. This is followed by an Exact deduplication method, which trims another 13 billion tokens.  Finally, the text cleaning process identifies and excludes 5B improper tokens, including all sensitive information. Ultimately, the pipeline crafts a streamlined financial dataset, culminating in a concise 20B-token financial dataset. The pipeline is illustrated in figure \ref{fig:dedup_pipeline}. 

\section{Downstream Dataset details}
\label{dds}
Notable datasets include FPB and FiQA-SA, both utilized for sentiment analysis, with the former comprising 48,450 news texts \citep{malo2013good} and the latter encompassing 11,730 news headlines and tweets \citep{inproceedings}. The FOMC dataset, consisting of 496 FOMC transcripts, serves the hawkish-dovish classification task \citep{shah-etal-2023-trillion}, whereas the Headline dataset, with 11,412 news headlines, aids in news headline classification \citep{sinha2020impact}. Named entity recognition is the focus of the NER and Finer-Ord datasets \citep{salinas-alvarado-etal-2015-domain, shah2023finer}. We brought in ECTSUM and EDTSUM \citep{mukherjee-etal-2022-ectsum, zhou-etal-2021-trade} for text summarisation. For text classification, we included two credit scoring datasets from the German and Australia  \citep{misc_statlog_(german_credit_data)_144, misc_statlog_(australian_credit_approval)_143}. We employed FinQA introduced by the current paper and ConvFinQA \citep{chen-etal-2021-finqa, chen-etal-2022-convfinqa} for numerical understanding task. We used three existing datasets for stock movement prediction, namely BigData22, ACL18, and CIKM18 \citep{soun2022accurate, xu2018stock, wu2018hybrid}. 

\noindent\textbf{Firm Disclosure Datasets}
This study employed three datasets that serve as a microcosm of firm regulatory disclosures. Each consists of labelled text segments from comprehensive reports annually filed with the Security and Exchange Commission (SEC) by public companies to inform investors regarding their financial health and business risks. The 'Firm Social Relationships' (FSR)  dataset provides insight into the intricate network of corporate interactions, categorized into several key relational dimensions: ownership, alliances, competition, and board interlock relationships \citep{caoetal2020}. They identified 3931 sentences stating another firm in a focal firm's disclosure. Domain experts classified the relationship between the focus firm and the firm into one or none of these relationships. The 'Cyber Strategies' (CS) dataset contains disclosure sentences describing the firm's cybersecurity strategies \citep{caoetal2023}. Experts labelled 240 cybersecurity-related sentences from firms' disclosures into one of five strategies delineated by the National Institute of Standards and Technology: Identification, Protection, Detection, Response, and Recovery \citep{NIST2018}. The 'IT Risk Disclosure' (ITR) dataset is created for this study using the Risk Factors section of the firm's annual disclosure. Domain experts categorized 1,196 sentences related to Information Technology into one or none of the 11 IT risk categories. These datasets curated by domain experts are pivotal to our zero-shot evaluation framework, which tests the models' utility against genuine instructional data—thus bridging the gap between theoretical model performance and practical utility in real-world scenarios.

\noindent\textbf{Financial Chart Understanding Dataset}
\begin{figure}[t]
\includegraphics[width=\columnwidth]{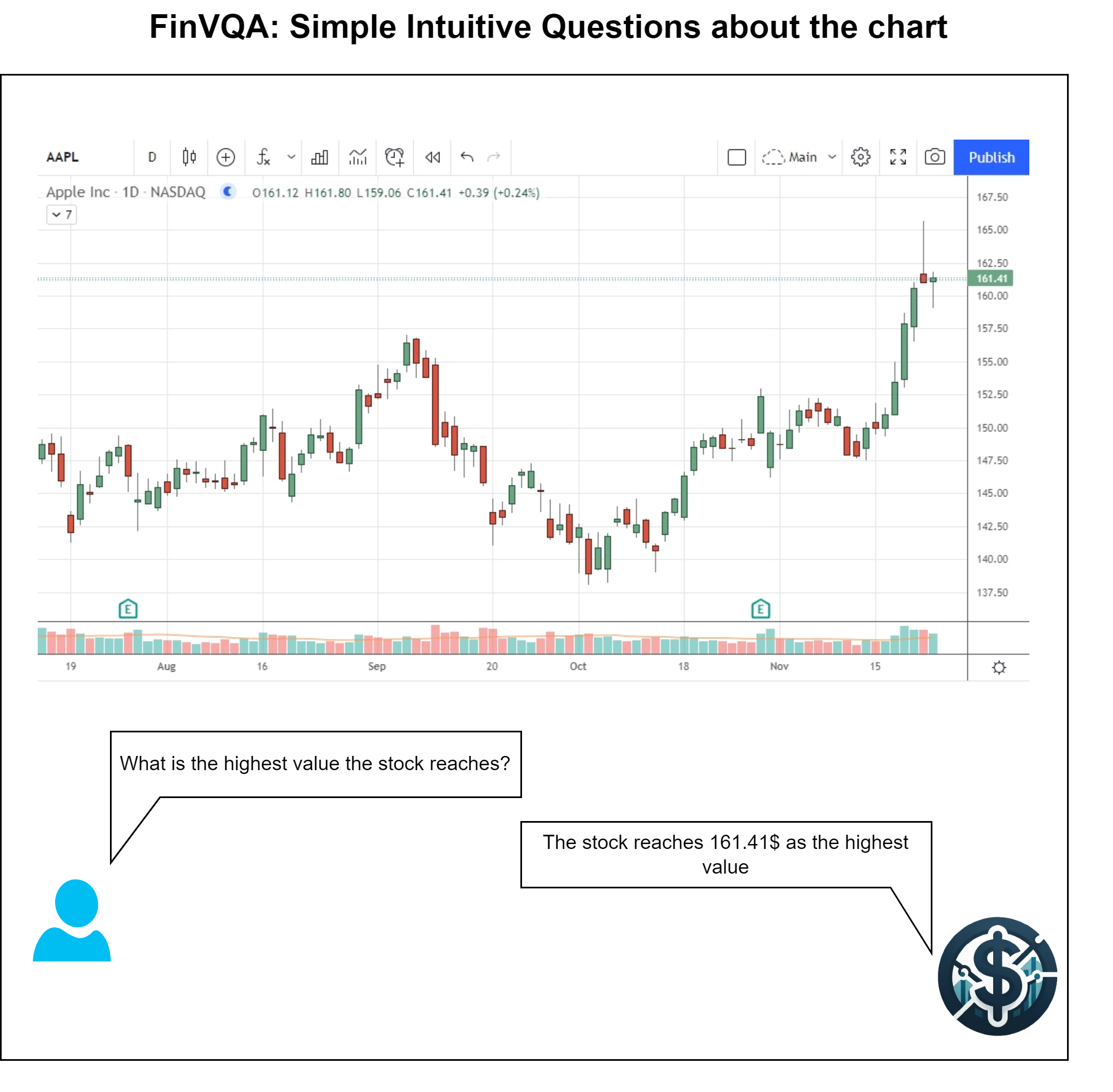}
\centering
\caption{Illustration of the FinVQA Dataset to understand model's proficiency in Chart Understanding Tasks. The figure demonstrates the model's ability to interpret stock price movements and identify peak values from a candlestick chart.} \label{fig:vl-fin}
\end{figure}

The FinVQA dataset addresses tasks involving questions about trends and details depicted in plots and graphs embedded in images. This dataset includes a variety of financial charts, such as line, bar, and candle charts
, all meticulously annotated by humans and accompanied by multi-turn conversations associated with each image. We developed two versions of the FinVQA dataset. The first, illustrated in Fig. \ref{fig:vl-fin}, focuses on simple questions related to stock market charts and requires the model to interpret these charts while identifying trends and performing analysis. These are randomly chosen 100 images that have been human annotated with different types of questions. 

\begin{figure}[t]
\includegraphics[width=\columnwidth]{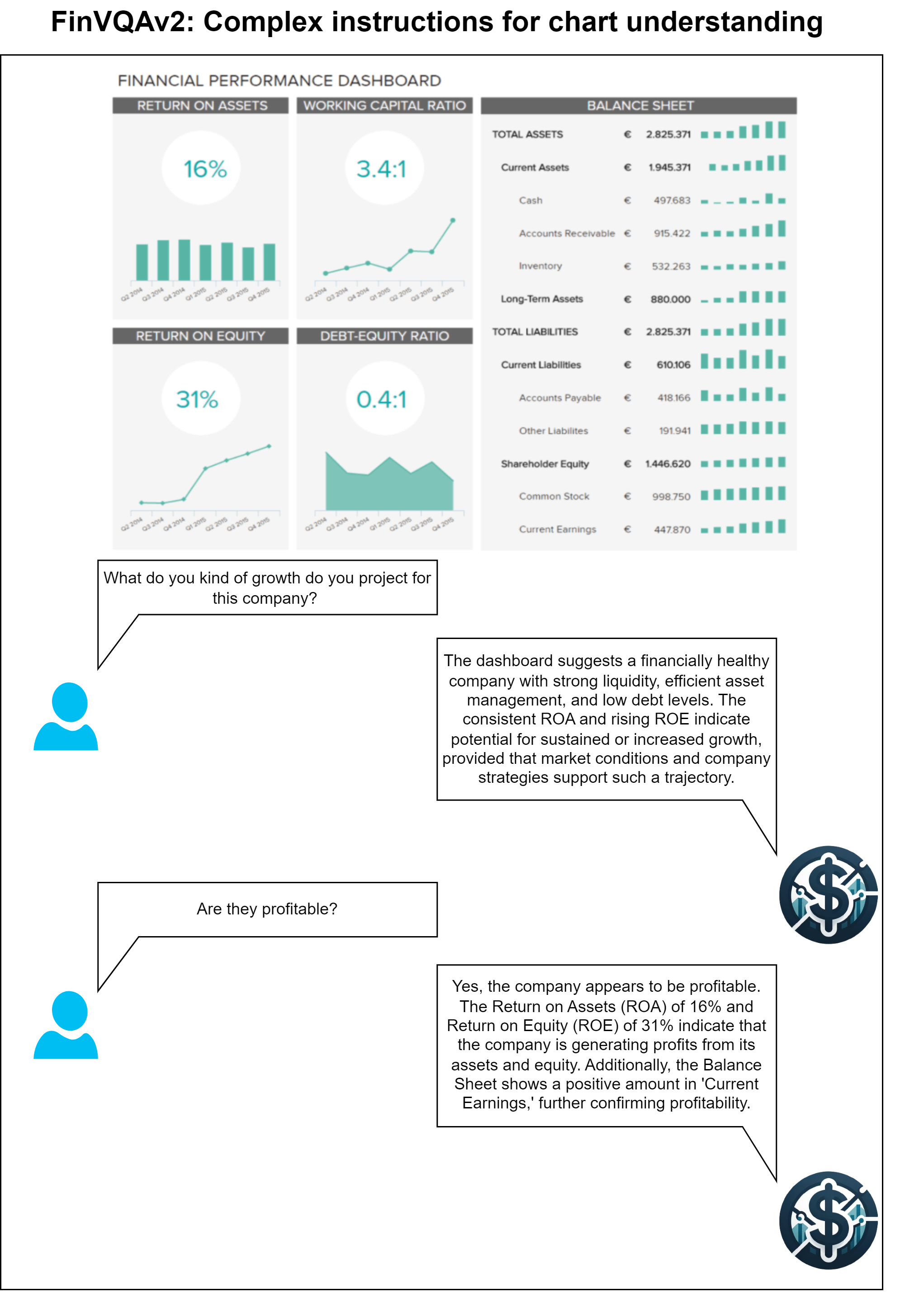}
\centering
\caption{Illustration of the FinBEAT VL Model's Proficiency in Chart Understanding Tasks. The panel showcases the model's adeptness at analyzing financial dashboards, projecting growth, and evaluating profitability based on key financial ratios and balance sheet data.} \label{fig:vl-fin-v2}
\end{figure}
FinVQAv2, presented in Fig. \ref{fig:vl-fin-v2}, is a dataset that captures more advanced and challenging situations. We gathered 100 images containing financial visualizations from various sources and collected relevant question-and-answer pairs from the experts to build our dataset. It encompasses a diverse array of financial graphics and various relevant questions posed in relation to each image. For instance, we present a financial dashboard containing various financial metrics in numerical and graphical formats, and we ask the model to perform complex calculations using the data extracted from the image. 

\noindent\textbf{Hallucinations Evaluation} In the financial context, Large Language Models (LLMs) like GPT-4 are prone to hallucinations, giving incorrect answers or misinterpreting basic facts, as shown by \citep{kang2023deficiency}. We generated two datasets: FinTerms-MCQ and FinTerms-Gen.

To build FinTerms-MCQ, we generated a dataset containing 1129 financial terms and their definitions, using the method described by \citep{ghosh-EtAl:2022:FNP}. This dataset assesses the foundational financial knowledge of various LLMs and investigates if retrieval-based methods can reduce the incidence of hallucinations. We built this dataset in a multiple-choice format with the question and four options; all four are closely related, and only one is correct. 

FinTerms-Gen is built as a generation task where we collected terms from \citet{Investopedia}, and then we asked our models to answer the definitions. Examples from this dataset are presented in Table \ref{tab:halo_eg}.

\begin{figure*}[!htb]
	\begin{center}
\begin{tikzpicture}[node distance=12mm]
	\node (a1) [heading] {\textbf{Retrieval Data extraction}};
	\node (a2) [heading, right= 5cm of a1] {\textbf{Answer Generation}};
	\node (n1) [prompt, node distance=5mm, below=of a1] {From the following PDF file, extract all the relevant information that might help in answering the question:\\ 
	\textbf{PDF:} \clb{\texttt{URL}} \\
        \textbf{Extracted Text:} \clvo{\textit{text........}} \\
	\textbf{Question:} \clb{\texttt{ask question?}} \\
 }; 
\node (n2) [prompt, node distance=5mm, below=of a2] 
	      {You are a \texttt{\clo{financial expert}} specializing in the nuanced analysis of financial statements and a wide array of data-driven financial tasks. And think step by step for each prompt. Sometimes, you must extract relevant information to proceed with the problem. \\
        \textbf{Instructions:} 
        \clm{\texttt{If any options are specified, ensure that your answer is one of the options specified. \\}}
		\phantom{new} \clm{\texttt{Do not explain why you think the answer is correct. \\} }

        \phantom{new} \cldg{\texttt{Answer the question by formulating your response using predefined mathematical functions. For addition, use Add(a, b), which represents a + b. Use Subtract(a, b) to denote a - b for subtraction. Construct your answer by combining these functions appropriately to reflect the required calculations.\\}}
        	\textbf{Context:} \clb{\texttt{text} + \texttt{table} + \texttt{image}} \\
        \textbf{Retrieval:} \clb{\texttt{Retrieved relevant information}} \\
	\textbf{Question:} \clb{\texttt{Task based question?}} \\
        \textbf{Answer:}
        };
	
\node (n4) [response, below=of n2]
              {\texttt{Program generated by the LLM.}};
\node (n3) [response, below=of n1]
	      {\texttt{Extracted text.}};
\node[above left=0.25cm and -3.50cm of n3] (image1) {\includegraphics[height=0.7cm]{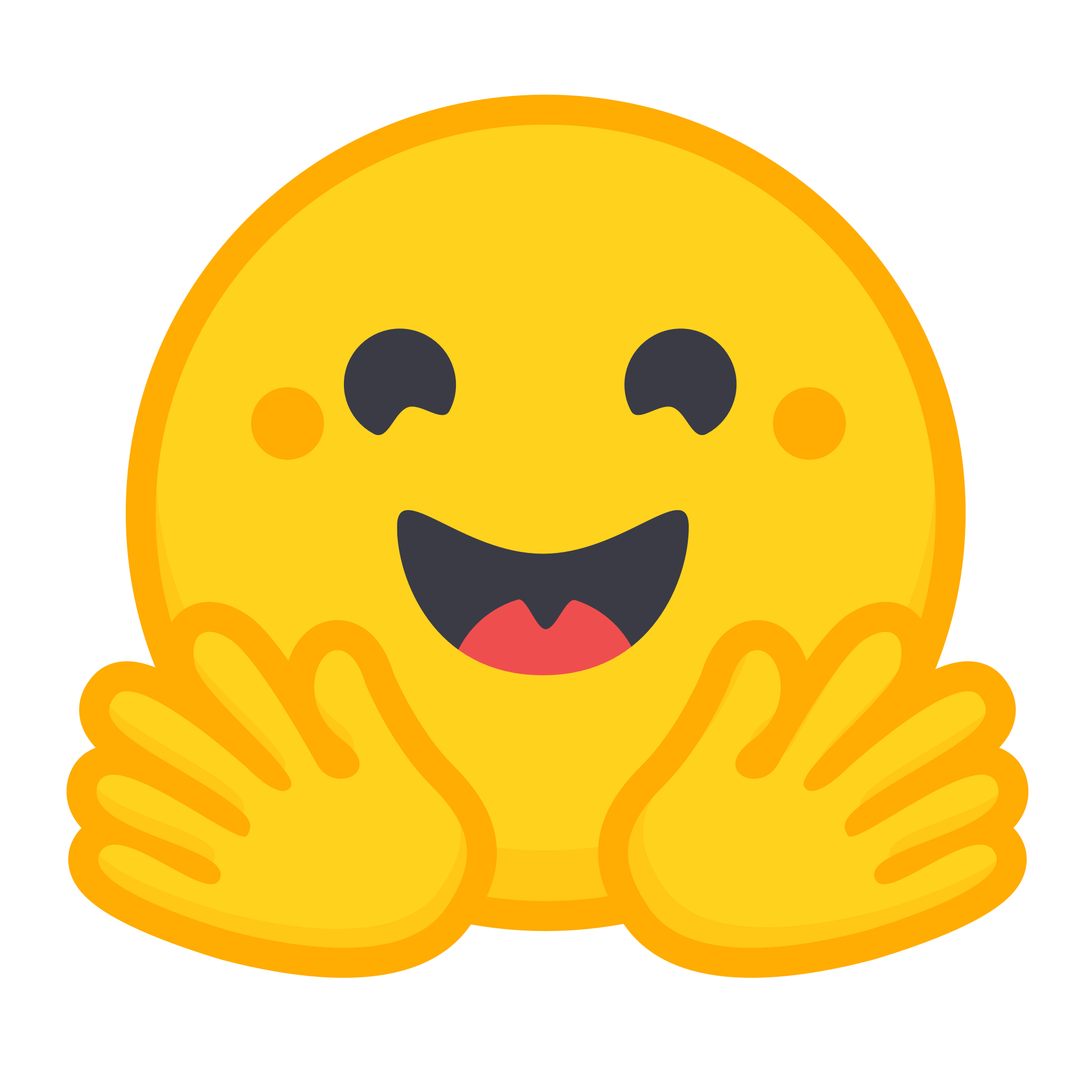}};
\node[below = -0.1cm of image1] {\scriptsize BGE};
\node[above left=0.25cm and -3.25cm of n4] (image2) {\includegraphics[height=0.7cm]{figures/logo_finbeat_big.png}};
\node[below= -0.25cm of image2] {\scriptsize FinTral};
\draw [arrow] (n1) edge[out=0, in=180, looseness=1.5] (n2);
\draw [arrow] (n3) edge[out=0, in=180, looseness=1.5] (n2);
\draw [arrow] (n1) -- (n3);
\draw [arrow] (n2) -- (n4);
	\node [fill=black, minimum width=0.05cm, minimum height=0.05cm, node distance=3mm, below left=0.25cm and 4.5cm of n4](c1) {};
	\node [right= -0.05cm of c1](cn1) {\scriptsize Signifier};
	\node [fill=orange, minimum width=0.05cm, minimum height=0.05cm, right= -0.01cm of cn1](c2) {};
	\node [right= -0.05cm of c2](cn2) {\scriptsize \clo{Memetic proxy}};
	\node [fill=magenta, minimum width=0.05cm, minimum height=0.05cm, right= -0.01cm of cn2](c3) {};
	\node [right= -0.05cm of c3](cn3) {\scriptsize \clm{Constraining behavior}};
	\node [fill=darkgreen, minimum width=0.05cm, minimum height=0.05cm, right= -0.01cm of cn3](c4) {};
	\node [right= -0.05cm of c4](cn4) {\scriptsize \cldg{Meta prompt}};
	\node [fill=blue, minimum width=0.05cm, minimum height=0.05cm, right= -0.01cm of cn4](c5) {};
	\node [right= -0.05cm of c5](cn5) {\scriptsize \clb{Input}};
\end{tikzpicture}
	\end{center}
\caption{Prompting Method for FinTral-RL-T\&R}
\label{fig:chain_of_ret}
\end{figure*}

\begin{figure*}[!htp]
	\begin{center}
\begin{tikzpicture}[node distance=12mm]
	\node (n1) [prompt] {You are a \texttt{\clp{financial expert}} specializing in the nuanced analysis of financial statements and a wide array of data-driven financial tasks. And think step by step for each prompt. Sometimes, you must extract relevant information to proceed with the problem. \\
        \textbf{Instructions:} \\
        \phantom{new} \clr{\texttt{If any options are specified, ensure that your answer is one of the options specified. \\}}
		\phantom{new} \clr{\texttt{Do not explain why you think the answer is correct. \\} }
  
        \textbf{Retrieval:} \clvo{Musk's December sale of Tesla stock worth \$3.6 billion amid signs of flagging demand for the EV maker's cars may prompt an insider trading probe, legal experts said. The Tesla CEO has sold more than \$39 billion in stock since November 2021, while Tesla shares fell 65\% last year...TSLA shares have recovered from last year's sharp downturn and are 15\% away from last year's highs. So far in 2023, Tesla's short sellers have seen roughly \$11.6 billion in mark-to-market losses. Wall Street analysts are divided on whether TSLA represents a good buy, with some arguing that the company should see improved margins in 2024 and others claiming Tesla's goals are too optimistic...} \\
	\textbf{Question:} \cllg{What is the sentiment of the following financial post: Positive, Negative, or Neutral?} \\
        \textbf{Context:} \clb{@scottrade Why is \$tsla not available for shorting at this time? Same for \$w?} \\
        \textbf{Answer:}
}; 

\node (n3) [response, below=of n1]
	      {\texttt{negative}};
\draw [arrow] (n1) --  (n3);
\node[above left=0.25cm and -3.25cm of n3] (image) {\includegraphics[height=0.9cm]{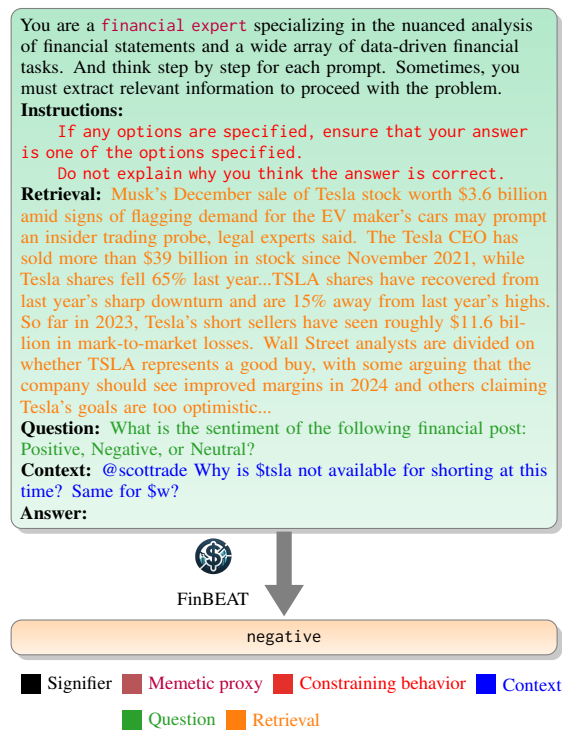}};
\node[below= -0.25cm of image] {\scriptsize FinBEAT};
	\node [fill=black, minimum width=0.05cm, minimum height=0.05cm, node distance=3mm, below left=0.25cm and -0.4cm of n3](c1) {};
	\node [right= -0.05cm of c1](cn1) {\scriptsize Signifier};
	\node [fill=cyan!05!purple, minimum width=0.05cm, minimum height=0.05cm, right= -0.01cm of cn1](c2) {};
	\node [right= -0.05cm of c2](cn2) {\scriptsize \clp{Memetic proxy}};
	\node [fill=cyan!05!red, minimum width=0.05cm, minimum height=0.05cm, right= -0.01cm of cn2](c3) {};
	\node [right= -0.05cm of c3](cn3) {\scriptsize \clr{Constraining behavior}};
        \node [fill=blue, minimum width=0.05cm, minimum height=0.05cm, right= -0.01cm of cn3](c4) {};
	\node [right= -0.05cm of c4](cn4) {\scriptsize \clb{Context}};
        \node [fill=lushgreen, minimum width=0.05cm, minimum height=0.05cm,right= -0.2cm of cn4, below = 0.2cm of c2](c5) {};
	\node [right= -0.05cm of c5](cn5) {\scriptsize \cllg{Question}};
	\node [fill=vibrantorange, minimum width=0.05cm, minimum height=0.05cm, right= -0.01cm of cn5](c6) {};
	\node [right= -0.05cm of c6](cn6) {\scriptsize \clvo{Retrieval}};
\end{tikzpicture}
	\end{center}
\caption{ FinTra Prompting method Example} 
\label{fig:FinBEAT_prompt_example}
\end{figure*}

\begin{figure*}[ht]
\includegraphics[width=\textwidth]{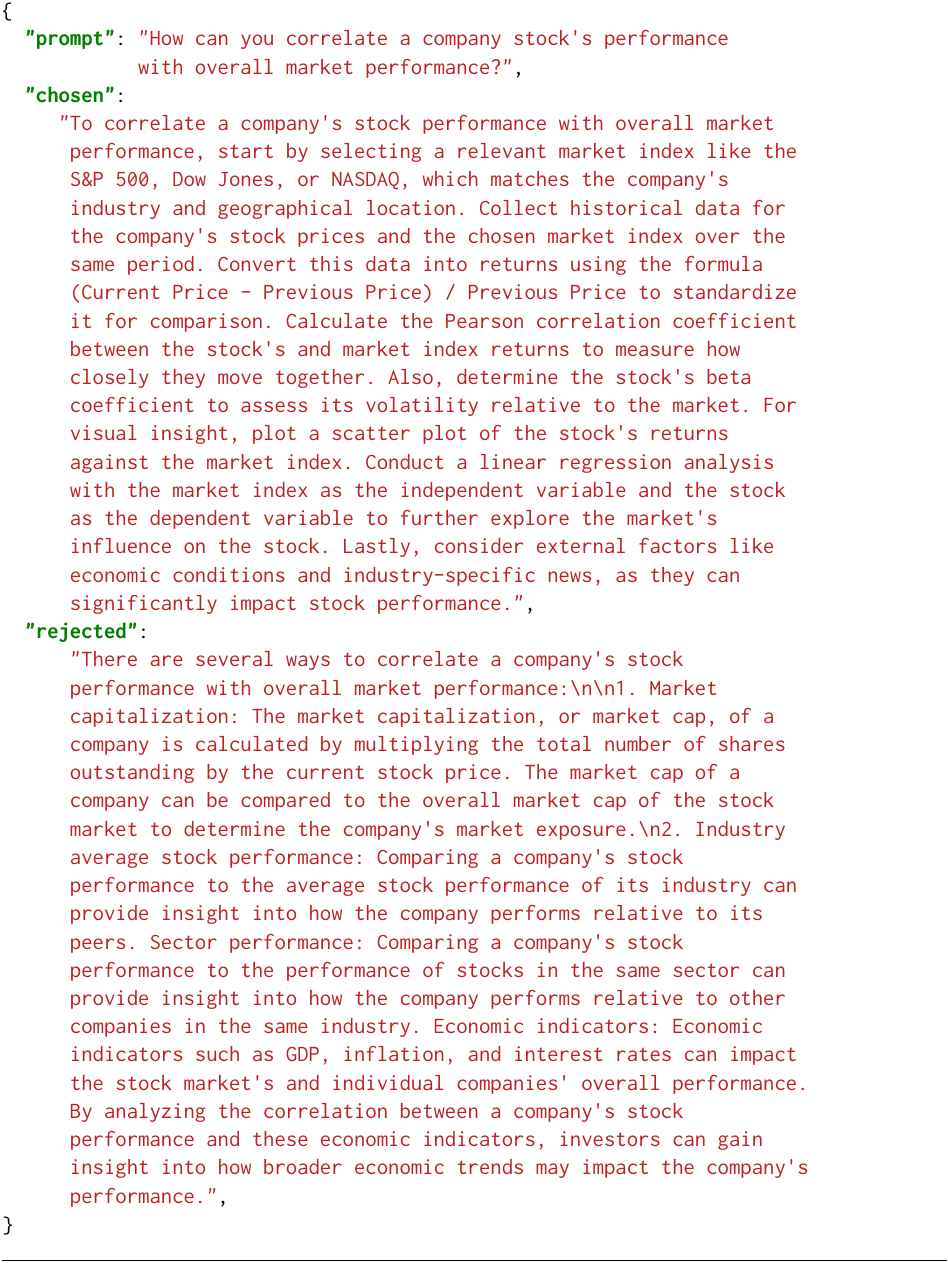}
\centering
\caption{Example from the Finance AI Feedback Dataset}
 \label{listing:dpo_data}
\end{figure*}

\begin{table*}[h]
    \centering
    \begin{tcolorbox}[colback=blue!5!white,colframe=blue!75!black,title=Hallucinations index]
        \begin{center}
            Examples of No Hallucinations 
        \end{center}
        \textbf{Prompt}: What is the definition of 'Low/No Documentation Loan' in the financial domain? \\
        \textbf{Wikipedia}: A No-Doc or Low-doc loan (abbr: No/Low Documentation Loan) refers to loans that do not require borrowers to provide income documentation to lenders or require little documentation. \\

        \textbf{FinTral-RL-T\&R}: A low/no documentation loan allows a 
        potential borrower to apply for a mortgage while providing little
        or no information regarding their employment, income, or assets. 
        Regulation of these loans has evolved significantly since 2008, 
        but they remain an option for some borrowers in nontraditional 
        financial situations. \\
        \textbf{Annotator Rating: \CircledA{A}} \\

        \textbf{ChatGPT}: A low/no documentation loan is a type of mortgage or loan where the borrower is not required to provide extensive financial documentation, such as income and asset verification, making it easier to qualify but often resulting in higher interest rates. \\
        \textbf{Annotator Rating: \CircledA{A}} \\

        \textbf{GPT-4}: A Low/No Documentation Loan is a type of loan that requires minimal or no verification of the borrower's income, assets, or employment, often used by individuals who may not have traditional income sources or documentation. \\
        \textbf{Annotator Rating: \CircledA{A}} \\
        
        \tcblower 
        \begin{center}
            Examples of LLMs Hallucinations 
        \end{center}
        \textbf{Prompt}: What is the definition of 'Open Market' in the financial domain? \\
        \textbf{Wikipedia}: An open market is an economic system without trade barriers to free market activities. In an open market, buyers and sellers can do business freely without common barriers, such as unfair licensing agreements, arbitrary taxes, unionization, subsidies, and other regulations affecting regular market operations.  \\
        
        \textbf{FinTral-RL-T\&R}: 	
        An open market is an economic system with little to no barriers to free-market activity. An open market is characterized by the absence of tariffs, taxes, licensing requirements, subsidies, unionization, and any other regulations or practices that interfere with free-market activity. Open markets may have competitive barriers to entry, but never any regulatory barriers to entry. \\
        \textbf{ Annotator Rating: \CircledA{A}} \\ 
        
        \textbf{ChatGPT}: In finance, the Open Market refers to a marketplace where securities are traded directly between buyers and sellers without the intervention of a third party, such as an exchange. \\
        \textbf{Annotator Rating: \CircledC{C} } \\
        
        \textbf{GPT-4}: An "Open Market" in the financial domain refers to an economic system with few barriers to free market activity. It typically involves unrestricted buying and selling of goods and services, with prices determined by supply and demand, and minimal government intervention in terms of tariffs, quotas, or regulations. This concept is fundamental in the operation of capital markets, where securities are freely traded. \\
        \textbf{Annotator Rating: \CircledA{A}} \\
        
    \end{tcolorbox}
    \caption{Examples of Hallucinations in LLMs}
    \label{tab:halo_eg}
\end{table*}\label{Appendix:halo_egg}

\begin{figure*}[ht]
\includegraphics[width=\textwidth]{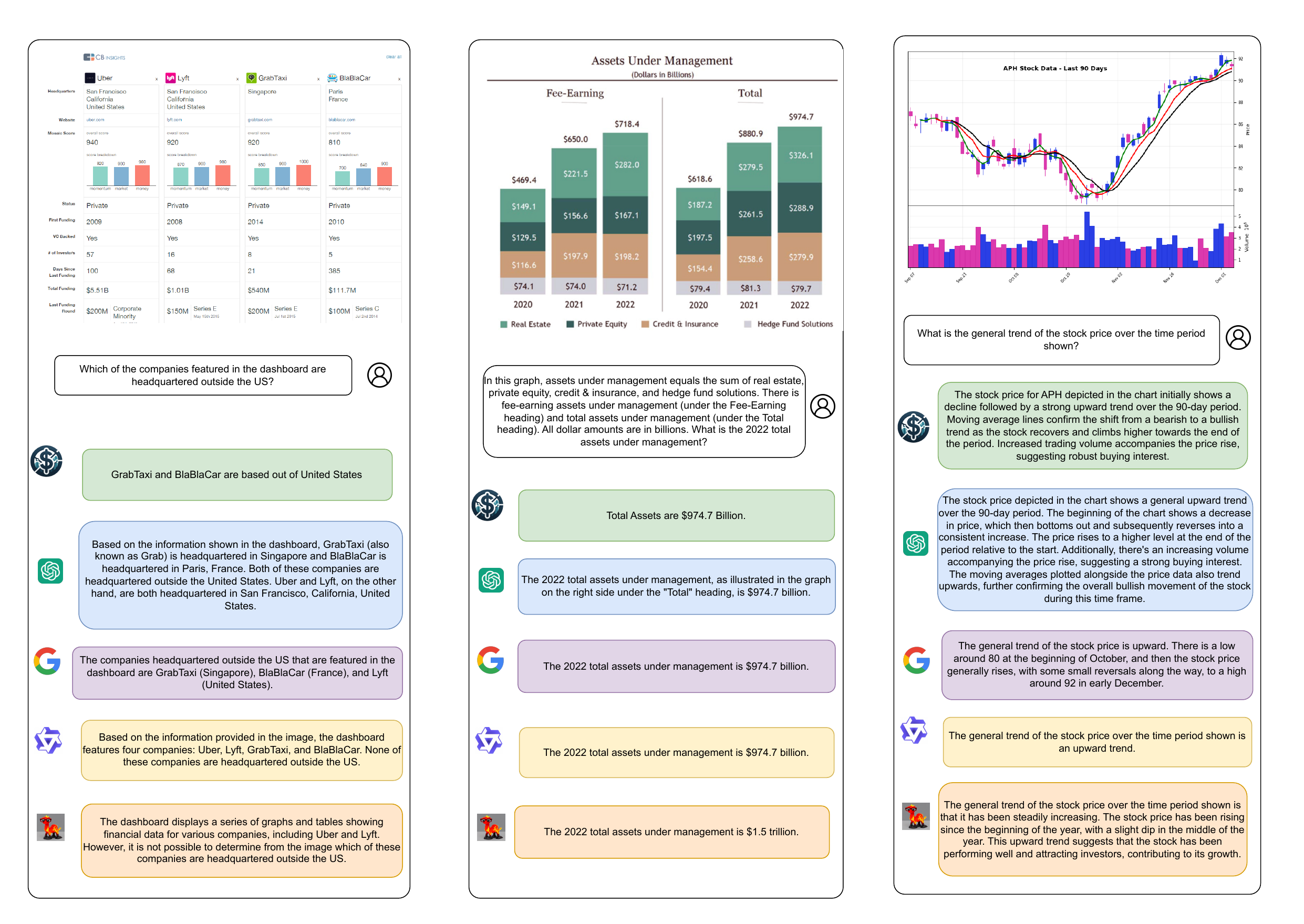}
\caption{Example of different VL models on different FinVQA tasks} \label{fig:vl-fin-fig}
\end{figure*}

\section{Applications} \label{Appendix:applications}
\begin{figure*}[ht]
\includegraphics[width=\textwidth]{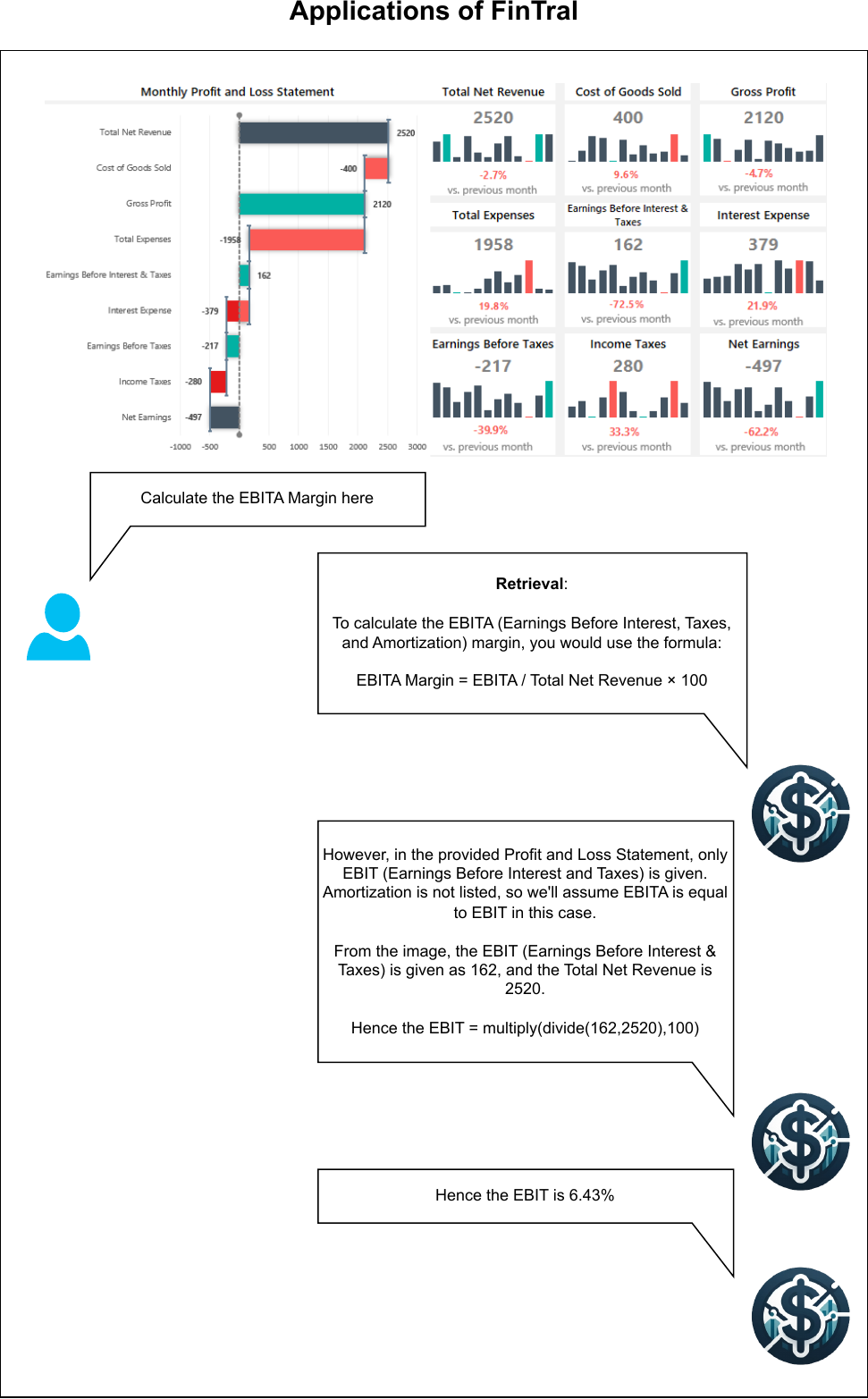}
\centering
\caption{Applications of the FinTral Model} \label{fig:applications}
\end{figure*}

Figure \ref{fig:applications} shows how this model can be used in real life. 
 \label{Appendix:applications}

\end{document}